\documentclass[pdflatex,sn-basic,Numbered]{sn-jnl}
\usepackage{graphicx}
\usepackage{tabularray}
\usepackage{multirow}
\usepackage{amsmath,amssymb,amsfonts}
\usepackage{amsthm}
\usepackage{mathrsfs}
\usepackage[title]{appendix}
\usepackage{xcolor}
\usepackage{textcomp}
\usepackage{manyfoot}
\usepackage{booktabs}
\usepackage{algorithm}
\usepackage{algorithmicx}
\usepackage{algpseudocode}
\usepackage{listings}
\usepackage{csquotes}
\usepackage{makecell}
\usepackage{caption}
\usepackage{enumitem}
\usepackage{array}
\usepackage[capitalise]{cleveref}
\usepackage[usestackEOL]{stackengine}
\usepackage{float}

\makeatletter
\newcolumntype{i}[1]{%
    >{\minipage[c]{\linewidth}\let\\\tabularnewline
      \itemize
      
      \addtolength{\rightskip}{0pt plus 1fil}
      \setlength{\itemsep}{-\parsep}}%
    p{#1}%
    <{\@finalstrut\@arstrutbox\enditemize\endminipage}}
\makeatother

\theoremstyle{thmstyleone}

\usepackage{threeparttable}
\usepackage{adjustbox} 
\theoremstyle{thmstyletwo}

\theoremstyle{thmstylethree}

\begin{document}

\title[Article Title]{CrownGen: Patient-customized Crown Generation via Point Diffusion Model}

\author[1]{\fnm{Juyoung} \sur{Bae}}\email{jbaeaa@cse.ust.hk}

\author[1]{\fnm{Moo Hyun} \sur{Son}}\email{mhson@cse.ust.hk}

\author[2]{\fnm{Jiale} \sur{Peng}}\email{u3012583@connect.hku.hk}

\author[2]{\fnm{Wanting} \sur{Qu}}\email{u3010894@connect.hku.hk}

\author[2]{\fnm{Wener} \sur{Chen}}\email{wechen22@connect.hku.hk}

\author[1]{\fnm{Zelin} \sur{Qiu}}\email{zqiuak@cse.ust.hk}

\author[7]{\fnm{Kaixin} \sur{Li}}\email{497155504@qq.com}

\author[7]{\fnm{Xiaojuan} \sur{Chen}}\email{1756833054@qq.com}

\author*[2]{\fnm{Yifan} \sur{Lin}}\email{yflin@hku.hk}

\author*[1,3,4,5,6]{\fnm{Hao} \sur{Chen}}\email{jhc@cse.ust.hk}

\affil[1]{\orgdiv{Department of Computer Science and Engineering}, \orgname{Hong Kong University of Science and Technology}, \orgaddress{\state{Hong Kong SAR}, \country{China}}}

\affil[2]{\orgdiv{Division of Pediatric Dentistry and Orthodontics, Faculty of Dentistry}, \orgname{University of Hong Kong}, \orgaddress{\state{Hong Kong SAR}, \country{China}}}

\affil[3]{\orgdiv{Department of Chemical and Biological Engineering}, \orgname{Hong Kong University of Science and Technology}, \orgaddress{\state{Hong Kong SAR}, \country{China}}}

\affil[4]{\orgdiv{Division of Life Science}, \orgname{Hong Kong University of Science and Technology}, \orgaddress{\state{Hong Kong SAR}, \country{China}}}

\affil[5]{\orgname{HKUST Shenzhen-Hong Kong Collaborative Innovation Research Institute}, \orgaddress{\state{Futian, Shenzhen}, \country{China}}}

\affil[6]{\orgdiv{State Key Laboratory of Nervous System Disorders}, \orgname{Hong Kong University of Science and Technology}, \orgaddress{\state{Hong Kong SAR}, \country{China}}}

\affil[7]{\orgname{Delun Dental Hospital}, \orgaddress{\state{Guangzhou}, \country{China}}}

\abstract{Digital crown design remains a labor-intensive bottleneck in restorative dentistry. We present \textbf{CrownGen}, a generative framework that automates patient-customized crown design using a denoising diffusion model on a novel tooth-level point cloud representation. The system employs two core components: a boundary prediction module to establish spatial priors and a diffusion-based generative module to synthesize high-fidelity morphology for multiple teeth in a single inference pass. We validated CrownGen through a quantitative benchmark on 496 external scans and a clinical study of 26 restoration cases. Results demonstrate that CrownGen surpasses state-of-the-art models in geometric fidelity and significantly reduces active design time. Clinical assessments by trained dentists confirmed that CrownGen-assisted crowns are statistically non-inferior in quality to those produced by expert technicians using manual workflows. By automating complex prosthetic modeling, CrownGen offers a scalable solution to lower costs, shorten turnaround times, and enhance patient access to high-quality dental care.}
\maketitle

\section{Introduction}\label{sec1}

The global burden of oral disease presents a persistent challenge to modern healthcare, with tooth loss
profoundly impacting masticatory function, facial aesthetics, and overall quality of life~\cite{peres2019oral, kassebaum2014global}. A crown's clinical success hinges on achieving a biomimetic morphology that harmonizes with the patient's unique occlusal scheme; failure can induce iatrogenic malocclusion and associated pathologies~\cite{fan2018occlusal}. Although computer-aided design/manufacturing (CAD/CAM) has digitized fabrication~\cite{miyazaki2009review}, the design phase remains an artisanal bottleneck~\cite{xie2025morphological, hosseinimanesh2025personalized}. Technicians must manually select and painstakingly adapt generic library templates to establish correct occlusal and proximal contacts—a process that can exceed an hour per crown~\cite{hosseinimanesh2025personalized,sailer2017randomized} and scales linearly for multiple restorations.

Generative artificial intelligence (GenAI) offers a transformative alternative to this paradigm. Early explorations reframed crown design as a 2D image synthesis task but were limited in scope and failed to capture full 3D anatomy~\cite{hwang2018learning, tian2021dcpr, tian2022dual}. Subsequent approaches using 3D voxel representations improved geometric completeness but struggled with the high resolutions needed for clinical fidelity, often sacrificing critical surface details~\cite{chau2024accuracy, ding2023morphology, farook2023computer}. More recent studies have converged on point cloud representations, which preserve sub-millimeter detail and represent the current state-of-the-art in single-crown generation~\cite{hosseinimanesh2025personalized, yang2024dcrownformer, lessard2022dental}.

Despite these advances, existing GenAI solutions are constrained by two fundamental limitations. First, they are invariably designed to generate a single crown. This is a direct consequence of an architecture that treats the entire dental arch as a monolithic geometric context. Trained on a fixed-input, fixed-output basis, these models lack any mechanism to parse multiple, arbitrarily located restoration sites and context teeth. This makes it clinically infeasible to train new models for every possible configuration of missing teeth. Second, these models are implicitly dependent on a prepared abutment tooth. The preparation's distinct geometry acts as a critical anchor, providing an unambiguous localization signal for the crown. In its absence—as in the planning stage of implant-supported restorations or bridge pontics—the model lacks the essential positional and scaling information to orient the prosthesis, rendering it incapable of generating a clinically acceptable result. This reliance restricts their use to traditional tooth-borne restorations and excludes a substantial portion of modern fixed interventions.

An ideal automated crown generation system must therefore satisfy three core requirements: it should (i) accept any arbitrary subset of existing teeth as conditional input, (ii) be able to synthesize any number of prostheses required, and (iii) remain agnostic to the presence or absence of abutment preparations to support the full spectrum of reduced tooth- and implant-borne restorations.

To address these gaps, we present CrownGen, a versatile framework that unites a denoising diffusion model~\cite{diffusion1ho, diffusion2sohldickstein, pvd} with a novel tooth-level representation. Our approach segments each tooth from the gingiva—a task robustly achievable with current models~\cite{xiong2023tsegformer, qiu2022darch, toothfairy1, li2024fine}—and treats each as an independent point cloud. This decomposition enables CrownGen to model the dentition not as a monolithic surface, but as a constellation of interacting objects. It learns inter-tooth relationships via our proposed Distance-weighted Inter-Tooth Attention (DITA) mechanism~\cite{shaw2018self, harvey2022flexible}, affording it the flexibility to generate one or many crowns in a single pass. Our evaluation demonstrates that CrownGen produces high-quality crowns that mimic natural morphology for restorations involving up to six missing teeth.

We validated CrownGen's performance through a two-protocol evaluation. Protocol 1 quantified its geometric fidelity against healthy, ground-truth natural teeth from 496 external test scans, demonstrating that CrownGen significantly outperformed three state-of-the-art methods across diverse restoration scenarios. Protocol 2 assessed its clinical utility on 26 restoration cases from 23 patients. Two trained clinical dentists compared crowns from a conventional, fully manual CAD workflow with those from our CrownGen-assisted workflow. Results show the CrownGen-assisted workflow significantly reduces design time while producing crowns of clinical quality statistically non-inferior to those from expert technicians, highlighting its potential to shorten design times, lower costs, and enhance patient access to high-quality dental care.

\begin{figure}[H]
\centering
\includegraphics[width=1.0\textwidth]{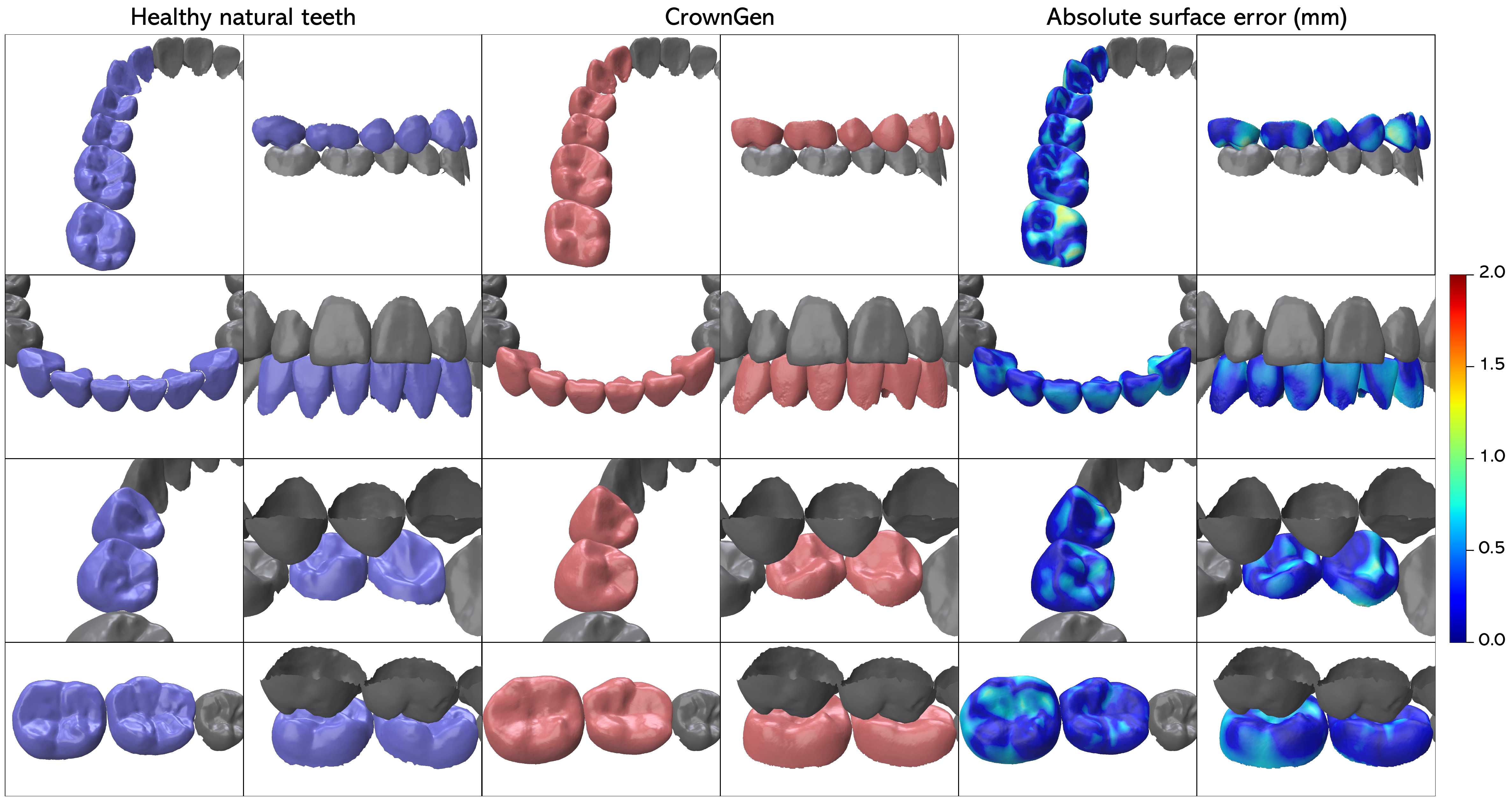}
\captionsetup{width=1.0\textwidth, skip=7pt} 
\caption{\textbf{Visual comparison between ground-truth natural teeth and crowns generated by CrownGen.} Each row corresponds to a simulated clinical case involving multiple missing teeth for restoration. From top to bottom, scenarios shown are the restoration of a maxillary posterior sextant (FDI 12–17), mandibular anterior teeth (FDI 33–43), adjacent mandibular premolars (FDI 34–35), and adjacent mandibular molars (FDI 36–37). The rightmost column presents a surface distance error map, visualizing the point-to-surface deviation between the generated crown and the ground-truth mesh. CrownGen consistently generates anatomically plausible crowns that closely emulate the morphology of the original, healthy dentition.}\label{fig_meshrecon_visual}
\end{figure}

\section{Results}\label{sec2}

\subsection{The CrownGen Framework}\label{sec2sub1}
CrownGen is an automated digital prosthodontics framework designed to generate patient-customized crowns for diverse and complex restorative scenarios. Its architecture is built upon two synergistic deep-learning modules that decouple the tasks of spatial localization and high-fidelity shape generation: a boundary prediction module and a point cloud diffusion-based generative module.

First, the boundary-prediction module addresses the critical initial step of determining the location, orientation, and scale for each required prosthesis. Conditioned on the set of context teeth $\mathcal{Y}$ and the FDI labels of the missing teeth (generation targets), the module predicts a tightly fitting cylindrical boundary for each target site. This design delegates the responsibility of resolving fine-grained details, such as subtle tooth rotations and precise morphology, to the subsequent generative module. Crucially, by decoupling localization from shape generation, the predicted boundary acts as a powerful spatial prior, effectively constraining the stochastic generation process within a high-probability region and allowing the model to allocate its full representational power to resolving intricate morphological features.

The core of our framework is the diffusion-based crown generation module, which synthesizes the final anatomical forms. Conditioned on the context teeth $\mathcal{Y}$ and the predicted cylindrical boundaries, the module generates a collection of crowns $\mathcal{X}$ through a learned stochastic denoising process~\cite{diffusion1ho, diffusion2sohldickstein}. The denoising network incorporates our proposed Distance-weighted Inter-Tooth Attention (DITA) layers. This mechanism enables the model to intelligently prioritize local context, allowing adjacent and antagonistic teeth to exert a stronger morphogenetic influence (Supplementary Figure~\ref{fig_dita}). The output is a set of discrete, high-resolution point clouds, each representing a fully formed crown. Finally, to ensure clinical applicability, each generated point cloud is transformed into a watertight mesh using a learnable Poisson surface reconstruction model~\cite{dpsr}, rendering it ready for downstream CAD/CAM workflows.

A key challenge in training such a versatile model is data heterogeneity. While fully dentate scans provide ideal training targets, clinical datasets are dominated by scans with pre-existing edentulous sites or damaged teeth, which are unsuitable for standard supervised learning as they lack proper contextual references. To overcome this, we adopted a two-stage pseudo-crown training strategy. In the first stage, we trained an initial version of the generative module exclusively on a smaller cohort of scans exhibiting a complete and healthy 28-tooth permanent dentition. In the second stage, we used this preliminarily trained module to perform inference on the larger, partially edentulous dataset, synthesizing anatomically plausible \enquote{pseudo-crowns} to fill the missing-teeth gaps. This data augmentation scheme allowed us to create a vast, complete, and anatomically consistent training set.

\subsection{Generation Performance on Natural Dentition}\label{sec2sub4}

To comprehensively evaluate CrownGen's generative capabilities, we benchmarked its performance against ground-truth natural teeth from an external test set of 496 healthy dentitions and compared it with three state-of-the-art point cloud completion methods: PointSea~\cite{pointsea}, AdaPoinTr~\cite{adapointr}, and ProxyFormer~\cite{proxyformer}. Geometric fidelity was assessed using established metrics at both the raw point cloud and final reconstructed mesh levels.

\subsubsection{Boundary Prediction Accuracy}\label{sec2sub4sub1}

\begin{table}[h]
\renewcommand{\arraystretch}{1.2}
\begin{tabular}{lcccc}
\toprule%
& \multicolumn{2}{@{}c}{Dice Coefficient}
& \multicolumn{2}{c@{}}{Intersection over Union} \\
\cmidrule(l){2-3}\cmidrule(r){4-5}
Tooth type&$\text{Mean}\pm\text{SD}$&95\% CI&$\text{Mean}\pm\text{SD}$ & 95\% CI\\
\midrule
All types $(n=16,368)$          & $0.883\pm0.041$ & $(0.882-0.884)$ & $0.796\pm0.061$ & $(0.795-0.797)$ \\
Central incisor $(n=2,347)$     & $0.897\pm0.032$ & $(0.896-0.899)$ & $0.817\pm0.051$ & $(0.815-0.819)$ \\
Lateral incisor $(n=2,277)$     & $0.886\pm0.039$ & $(0.885-0.888)$ & $0.801\pm0.059$ & $(0.798-0.803)$ \\
Canine $(n=2,392)$              & $0.883\pm0.038$ & $(0.881-0.885)$ & $0.795\pm0.056$ & $(0.793-0.798)$ \\
First premolar $(n=2,335)$      & $0.885\pm0.040$ & $(0.883-0.887)$ & $0.799\pm0.057$ & $(0.796-0.801)$ \\
Second premolar $(n=2,333)$     & $0.879\pm0.043$ & $(0.877-0.881)$ & $0.789\pm0.059$ & $(0.786-0.791)$ \\
First molar $(n=2,355)$         & $0.891\pm0.036$ & $(0.890-0.893)$ & $0.808\pm0.057$ & $(0.806-0.810)$ \\
Second molar $(n=2,329)$         & $0.859\pm0.048$ & $(0.857-0.861)$ & $0.761\pm0.069$ & $(0.758-0.763)$ \\
\botrule
\end{tabular}
\captionsetup{skip=7pt} 
\caption{\textbf{Boundary prediction accuracy for crown localization.} Performance of the boundary prediction module evaluated on $n = 16,368$ individual tooth boundaries derived from 496 test dentitions, with 95\% confidence intervals estimated via a nonparametric bootstrap with $10,000$ resamples.}\label{tab_bound}
\end{table}

We first evaluated the performance of the boundary prediction module, as its accuracy is critical for the subsequent generative process. This analysis utilized 6,944 partially edentulous test scenario dentitions that involve 16,368 individual target boundaries. We quantified accuracy by comparing predicted cylindrical boundaries to their ground-truth using the Dice Coefficient and Intersection over Union (IoU). The module demonstrated high predictive accuracy across all tooth types, achieving a mean Dice of 0.883 and IoU of 0.796 (Table~\ref{tab_bound}). Crucially, this performance was highly consistent across the dental arch, with the mean Dice ranging from a high of 0.897 for central incisors to 0.859 for the anatomically variable second molars. These results confirm that the module provides a reliable and precise spatial prior for the generative stage, irrespective of the target tooth's location or morphology.

\begin{figure}[H]
\centering
\includegraphics[width=1.0\textwidth]{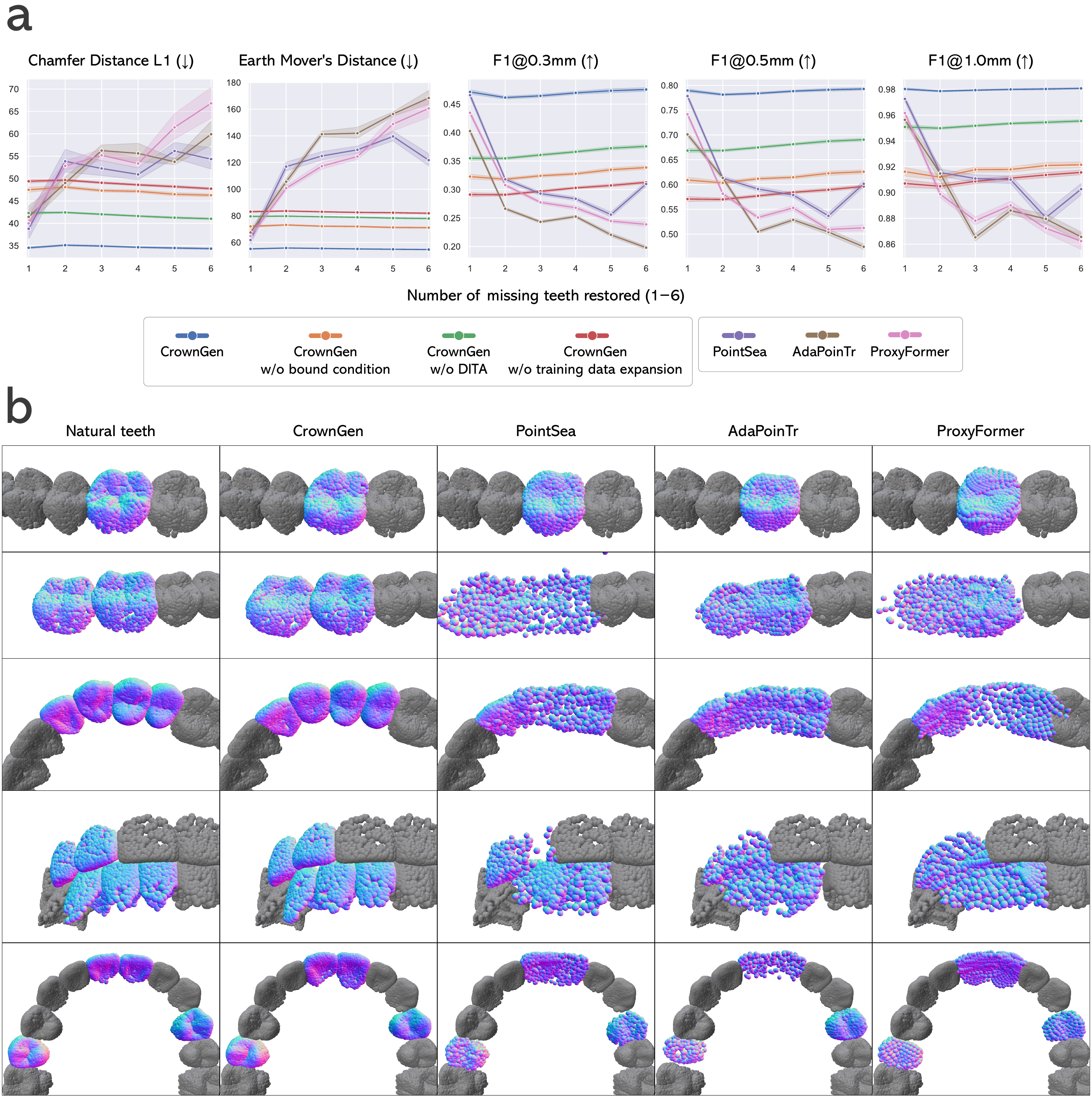}
\captionsetup{width=1.0\textwidth, skip=7pt} 
\caption{
    \textbf{Geometric fidelity of generated point cloud crowns.}
    \textbf{a.} Mean performance (curves) with 95\% CIs (shaded bands; nonparametric bootstrap, $10,000$ resamples) of CrownGen, its ablated variants, and state-of-the-art methods across scenarios constructed from 496 test dentitions. Performance is aggregated across all scenarios, encompassing all tooth types and stratified by the number of missing teeth restored. Metrics annotated with ↓ indicate lower is better, and vice versa. Chamfer distance and Earth mover’s distance are scaled by $10^{3}$. All pairwise differences relative to CrownGen are significant (two-sided paired $t$-test; $p<0.01$). Performance stratified by tooth functional group is shown in Supplementary Figure~\ref{fig_pointcloud_toothgroup}. Sample sizes for each condition are detailed in Supplementary Table~\ref{tab_pointcloud_all}.
    \textbf{b.} Visual comparison of generated point clouds for five representative restoration scenarios (top to bottom): single posterior molar (FDI 16); adjacent mandibular molars (FDI 46, 47); contiguous anterior–premolar group (FDI 22–25); bilateral, inter-arch restoration (FDI 13, 43, 12, 42, 41); and four anatomically dispersed teeth (FDI 15, 11, 21, 24). While comparing methods yield plausible single-tooth shapes, they fail to produce distinct, anatomically coherent structures in multi-tooth settings. In contrast, CrownGen consistently generates discrete, well-formed crowns that respect individual tooth morphology.
}
\label{fig_pointcloud_all}
\end{figure}

\subsubsection{Performance Comparison with State-of-the-art Methods}\label{sec2sub4sub2}
We next compared CrownGen against the three state-of-the-art point cloud completion methods: PointSea~\cite{pointsea}, AdaPoinTr~\cite{adapointr}, and ProxyFormer~\cite{proxyformer}. The quantitative results, detailed in Figure~\ref{fig_pointcloud_all}a and Supplementary Figure~\ref{fig_pointcloud_toothgroup}, reveal that CrownGen consistently and significantly outperforms all competing methods across all evaluation metrics and for every number of missing teeth restored. While all methods delivered reasonable performance in the single-tooth restoration scenario, their efficacy collapsed as the complexity of the restoration increased.

Specifically, when the number of restorations required per scenario exceeded one, the performance of the comparing methods dropped precipitously, accompanied by high variance in their results. This degradation is an inherent consequence of their architectural paradigm. By processing the entire context dentition as a single, monolithic point cloud, these models struggle to parse the complex input and generate multiple, anatomically distinct structures in arbitrary locations. As the number and spatial dispersion of missing teeth increase, the completion task becomes combinatorially more complex as missing teeth can appear at any location in the dentition~\cite{liu2018image, han2017high}, leading to a marked drop in fidelity.

In stark contrast, CrownGen's tooth-level object representation and explicit inter-tooth attention mechanism circumvent this fundamental limitation. By modeling the dentition as a collection of interacting teeth, the framework's difficulty does not scale with the number of missing teeth being restored. This architectural advantage is reflected in its stable, high-quality performance across all scenarios. For some metrics, CrownGen's performance exhibited a slight improvement as the number of missing teeth restored increased. We speculate that a larger and more globally distributed set of missing teeth compels the model to integrate information from a wider range of context teeth, providing richer, more robust conditional guidance compared to scenarios where generation is highly localized.

These quantitative findings are corroborated by qualitative visual inspection as shown in Figure~\ref{fig_pointcloud_all}b. While all comparing methods produce plausible results for a single-tooth restoration, they fail to maintain coherent tooth morphology in multi-tooth restoration scenarios, often generating an amorphous point cloud that merely \enquote{fills} the dentition without respecting individual anatomical forms. This lack of structural integrity renders their output unusable for downstream clinical applications. Points frequently appear in spurious locations beyond the crown boundaries, and the generated point clouds cannot be reliably segmented into individual crowns or reconstructed into watertight meshes—a non-negotiable step for per-tooth adjustment and fabrication.

\subsubsection{Ablation Study}\label{sec2sub4sub3}

\begin{figure}[h]
\centering
\includegraphics[width=0.85\textwidth]{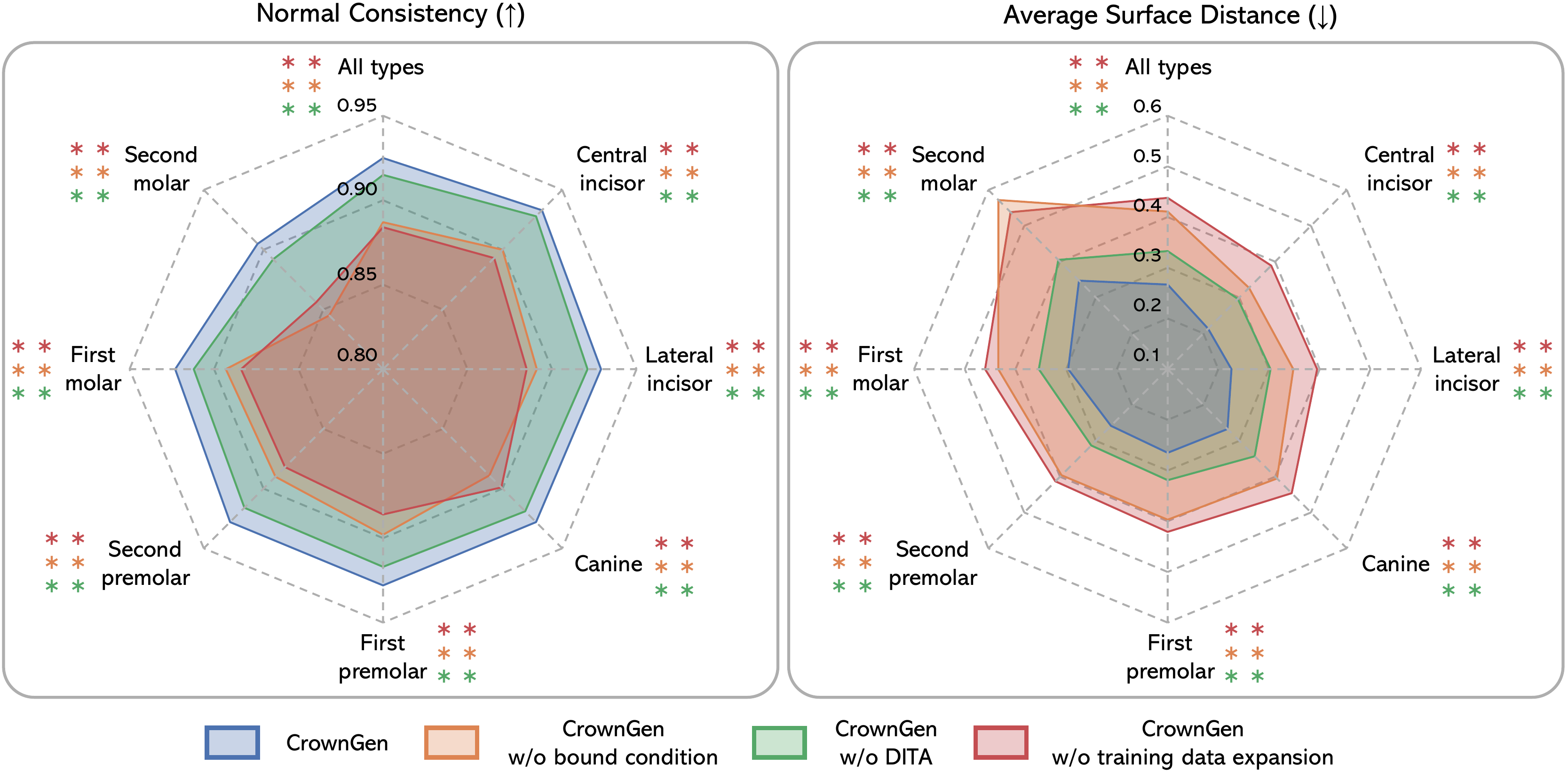}
\captionsetup{width=1.0\textwidth, skip=7pt} 
\caption{\textbf{Ablation study on final reconstructed mesh quality.} Performance comparison of CrownGen against its three ablated variants on the final reconstructed mesh surfaces of individual restored teeth across scenarios derived from 496 test dentitions. Metrics marked with ↓ indicate lower is better, and vice versa. Statistical differences relative to CrownGen were assessed using two-sided paired $t$-tests ($^{\ast\ast} p < 0.01$). Detailed sample sizes are provided in Supplementary Table~\ref{tab_ablation_metric}.}\label{fig_ablation_metric}
\end{figure}

To dissect the individual contribution of each key component of CrownGen, we performed an extensive ablation study. We evaluated three variants of our model: (i) CrownGen without the boundary prediction module, forcing the generative module to operate without an explicit spatial prior; (ii) CrownGen without the DITA layers, removing explicit inter-tooth attention and requiring the network to learn these relationships implicitly; and (iii) CrownGen trained without our pseudo-crown data expansion scheme, restricting the training set to only the complete, fully dentate scans. As all CrownGen variants generate discrete, well-confined, individual crowns, we were able to extend this analysis to include a post-reconstruction evaluation of the final mesh quality.

Our ablation study confirmed that each component of CrownGen is critical for high-fidelity crown generation. When evaluating the aggregate performance across all tooth types in the point cloud domain (Figure~\ref{fig_pointcloud_all}a, Supplementary Table~\ref{tab_pointcloud_all}), the fully-equipped CrownGen substantially outperformed all ablated variants. The performance gains were consistent and significant across all metrics. For Chamfer Distance $L1$ (CD), the full CrownGen outperformed its variants without the boundary module, DITA, and expanded data by $\sim$27\%, $\sim$17\%, and $\sim$29\%, respectively. This trend was amplified in stricter metrics: for Earth Mover's Distance (EMD), the corresponding gains were $\sim$23\%, $\sim$30\%, and $\sim$33\%, while for the $F1$ score (0.3 mm), they reached $\sim$43\%, $\sim$30\%, and $\sim$57\%.

This synergistic benefit was equally evident in the final reconstructed mesh quality, when evaluated across all test conditions (Figure~\ref{fig_ablation_metric}, Supplementary Table~\ref{tab_ablation_metric}). The complete CrownGen model improved Average Surface Distance (ASD) by 35.04\%, 19.82\%, and 39.04\% compared to models without the boundary module, DITA, and expanded data, respectively. Normal Consistency (NC) showed a similar pattern of improvement, with respective gains of 4.28\%, 1.09\%, and 4.63\%.

These substantial improvements highlight the crucial importance of boundary guidance, inter-tooth contextualization, and the scalability facilitated by our data augmentation strategy. The most severe performance degradation, particularly in CD, $F1$, and ASD metrics, resulted from ablating the training data expansion. This underscores the profound benefit and scalability of our self-bootstrapping strategy, which effectively leverages large-scale, partially edentulous clinical data. Note that the quality of the pseudo-crowns used for the training data expansion scheme is reflected by the performance of this variant trained without the expanded training data. While these pseudo crowns possess a lower geometric fidelity than those from the fully equipped variant of CrownGen, they still serve as robust anatomical placeholders. In these augmented scans, the contextual learning signal is overwhelmingly dominated by the numerous high-fidelity ICHnatural teeth, making the training process highly robust to the finer-grained inaccuracies of the few synthesized crowns.

Our analysis also revealed that the absence of the boundary prediction module degraded performance nearly as much as training only on complete dentitions, signifying the critical importance of providing an explicit spatial prior to guide and constrain the generative process. 
The ablation of inter-tooth attention layers was uniquely detrimental to the EMD metric, which measures global geometric correspondence. The pronounced impact on a globally sensitive, mass-preserving metric like EMD validates our assumption that explicit inter-tooth attention is crucial for capturing these global morphogenetic signals from adjacent and antagonistic teeth.

\subsection{Clinical Performance Evaluation}\label{sec2sub5}

\begin{figure}[H]
\centering
\includegraphics[width=1.0\textwidth]{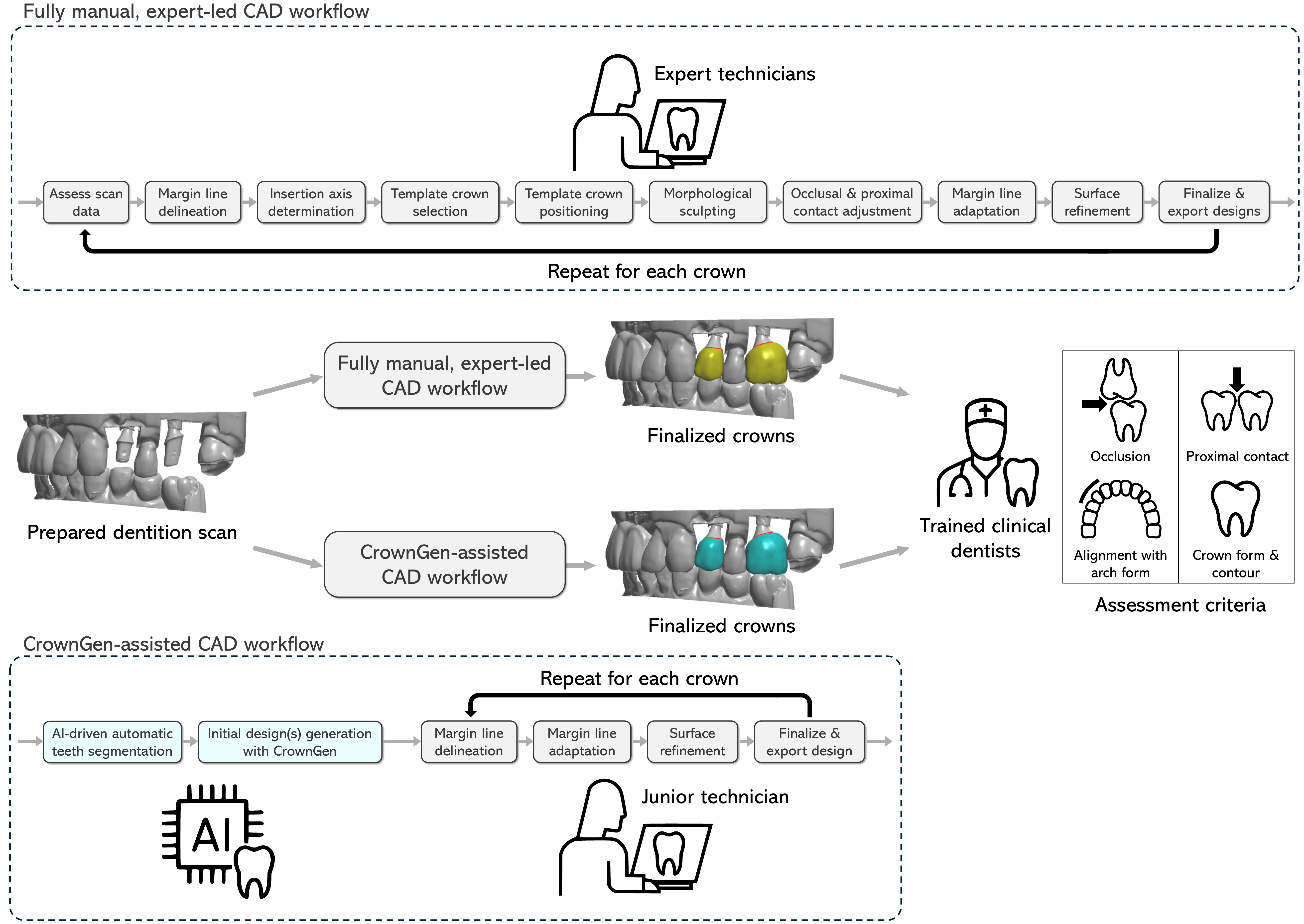}
\captionsetup{width=1.0\textwidth, skip=7pt} 
\caption{\textbf{Schematic of the clinical reader study design.} For each restoration case, a prepared dental scan was used to generate a digital crown via two parallel workflows: the conventional, fully manual CAD workflow and the CrownGen-assisted CAD workflow. The resulting paired designs were then presented to two trained clinical dentists for a comparative quality assessment.}\label{fig_reader_workflow}
\end{figure}

To assess the real-world utility of CrownGen, we conducted a formal reader study comparing crown designs from our CrownGen-assisted CAD workflow against those from a conventional, fully manual CAD workflow. This clinical evaluation was designed to determine if the CrownGen-assisted approach is (1) significantly more time-efficient and (2) statistically non-inferior in clinical quality.

\begin{figure}[H]
  \centering
\includegraphics[width=0.9\textwidth]{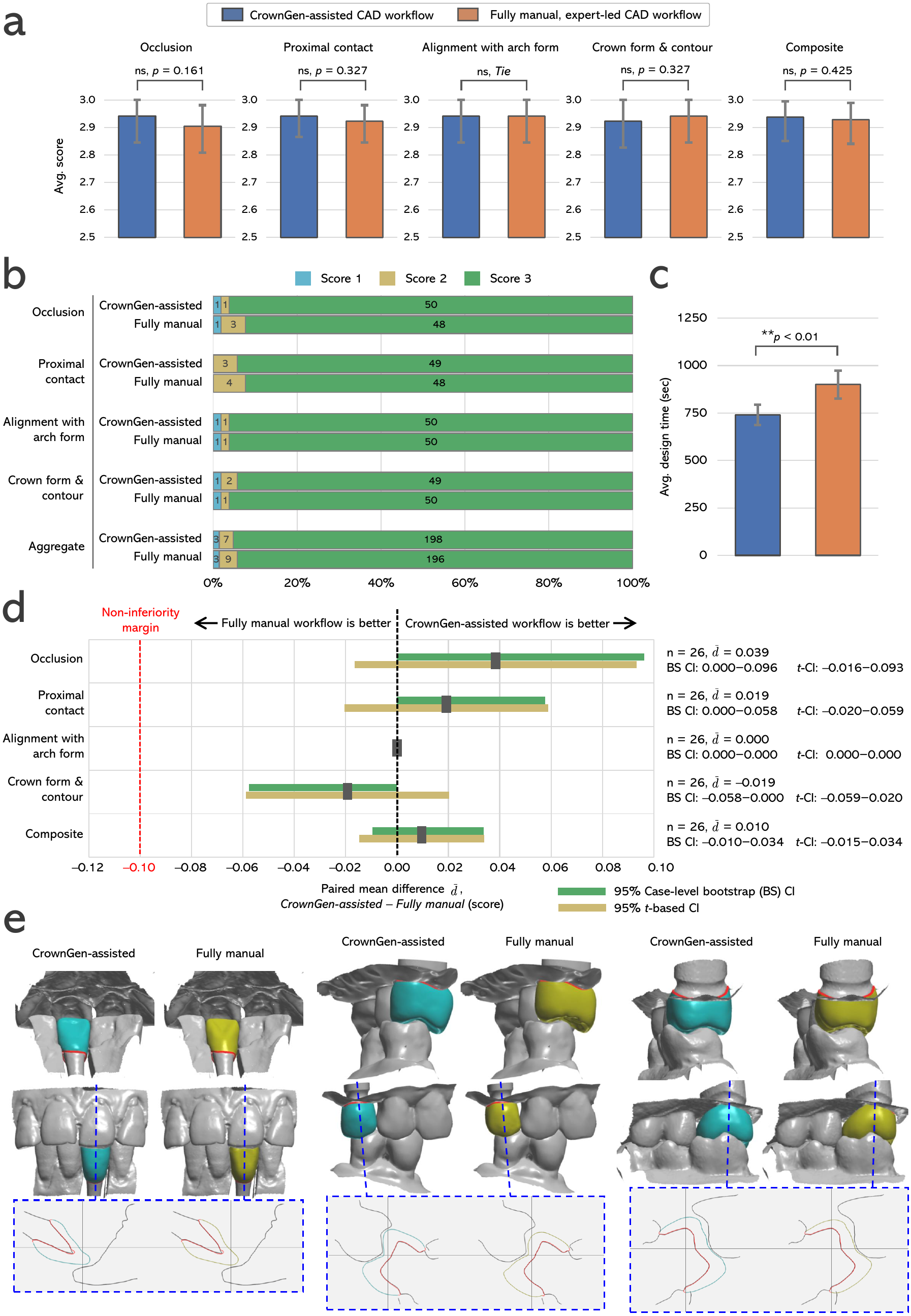}
\end{figure}
\begin{figure}[H]
\caption{
	\textbf{Overview of reader study results.} 
	\textbf{a.} Case-level average scores $(n=26)$ by workflow for each criterion and composite endpoints. For each case, the two readers’ scores were averaged per criterion; the composite is the per-case mean across the four criteria. Error bars show 95\% CIs (case-level, nonparametric bootstrap, $10,000$ resamples). Between-workflow differences were tested with a two-sided paired $t$-test.
	\textbf{b.} Stacked probability bars of ordinal scores (1–3) by criterion $\times$ workflow; each criterion bar aggregates 52 scores (26 cases $\times$ 2 readers). The aggregate endpoint pools all criteria (208 scores per workflow). 
	\textbf{c.} Design time by workflow (mean across cases). Error bars show 95\% CIs. A one-sided paired $t$-test assesses faster design time ($^{\ast\ast}p<0.01$) of CrownGen-assisted CAD workflow over a fully manual, expert-led CAD workflow.
    \textbf{d.} Forest plot of paired mean differences in score by criterion and composite. Points mark the sample paired mean difference $\bar d$; whiskers show 95\% CIs from nonparametric, case-level bootstrap CI (primary) and parametric paired $t$-based CI (sensitivity). The pre-specified non-inferiority margin is overlaid; CIs lying entirely above this margin indicate non-inferiority of the CrownGen-assisted CAD workflow compared to the fully manual, expert-led CAD workflow.
    \textbf{e.} Visual comparison of the patients' initial dentitions with the final digital crown designs generated by the CrownGen-assisted and fully manual workflows. The top two rows present external views, and the bottom row provides a cross-sectional view to illustrate internal adaptation and the occlusal contact scheme. Visualizations were created using the 3Shape 3D Viewer (version 1.3).
}\label{fig_reader_result_3point}
\end{figure}

\subsubsection{Workflow Efficiency}
We first compared the active design time required for each workflow across 26 crown restoration cases. The CrownGen-assisted CAD workflow demonstrated a statistically significant reduction in design time, requiring a mean of 740$\pm$131 seconds (95\% CI: 686–794 s). In contrast, the fully manual CAD workflow by highly skilled technicians required a mean of 900$\pm$180 seconds (95\% CI: 825–974 s) (Figure~\ref{fig_reader_result_3point}c). This represents a 17.78\% reduction in active design time ($p < 0.01$, one-sided paired $t$-test), 
underscoring the efficiency gains offered by our framework.

\subsubsection{Clinical Quality Assessment}
The clinical quality of crowns from both workflows was evaluated by two independent readers with a Likert rating scale of 1 (worst) through 3 (best), spanning across four key functional criteria: Occlusion; Proximal contact; Alignment with arch form; and Crown form and contour (Supplementary Table~\ref{tab_rubric}). Out of total 208 aggregated ratings (26 cases $\times$ 2 readers $\times$ 4 criteria) for each workflow, crowns from the CrownGen-assisted workflow were rated as clinically acceptable (score of 3) in 95.2\% of instances (198/208), compared to 94.2\% for the fully manual workflow (196/208), indicating high clinical utility for both methods (Figure~\ref{fig_reader_result_3point}b).

The primary endpoint for clinical quality was a composite score calculated for each of the 26 cases, representing the mean of the four reader-averaged criterion scores. A formal statistical comparison of these composite scores revealed no significant difference between the two workflows. The mean composite score was 2.938 (95\% CI: 2.851–2.995) for the CrownGen-assisted workflow and 2.928 (95\% CI: 2.841–2.990) for the fully manual workflow ($p = 0.425$, two-sided paired $t$-test). This finding of equivalence extended to the individual criteria, where no statistically significant differences were observed for Occlusion ($p = 0.161$); Proximal contact ($p = 0.327$); or Crown form and contour ($p = 0.327$). Notably, for the Alignment with arch form criterion, both workflows received identical mean scores (Figure~\ref{fig_reader_result_3point}a, Supplementary Table~\ref{tab_reader_summary}).

\subsubsection{Non-inferiority Analysis}
To formally test whether the CrownGen-assisted workflow produces crowns of at least comparable quality, we performed a non-inferiority (NI) analysis on the paired mean difference in reader scores. Based on established precedents in prosthodontic literature, we prespecified a conservative NI margin of $-0.10$ points on the 3-point rating scale, which corresponds to a 5-percentage-point margin. NI is established if the lower bound of the 95\% confidence interval (CI) for the mean paired difference in scores (CrownGen-assisted$-$Fully manual) remains above this margin. We computed two types of CIs: a case-level bootstrap CI (primary analysis) used to determine NI, and a parametric $t$-based CI (sensitivity) reported for robustness.

For the composite endpoint, paired differences favored CrownGen; however, lower confidence limits crossed zero, so superiority was not established. Crucially, for the composite and for Occlusion and Proximal contact, the lower bounds of two-sided 95\% CIs from the primary case-level bootstrap and the sensitivity $t$-based interval both exceeded the NI margin; NI was therefore concluded. For Alignment with arch form, all case-level paired differences were identically zero, yielding a zero-width interval that still satisfies NI. For Crown form and contour, point estimates slightly favored the fully manual workflow, yet the lower bounds of two-sided 95\% CIs remained above the NI margin; NI was again concluded. Notably, for Occlusion and Proximal contact, the (primary) bootstrap lower bounds were exactly 0, reflecting many ties in reader-averaged, discrete scores. Consequently, we conclude that the CrownGen-assisted CAD workflow is non-inferior to the conventional, fully manual CAD workflow by expert technicians in producing clinically high-quality dental crowns. The forest plot summarizes paired differences and CI lower bounds across endpoints (Figure~\ref{fig_reader_result_3point}d).

\begin{figure}[H]
\centering
\includegraphics[width=0.9\textwidth]{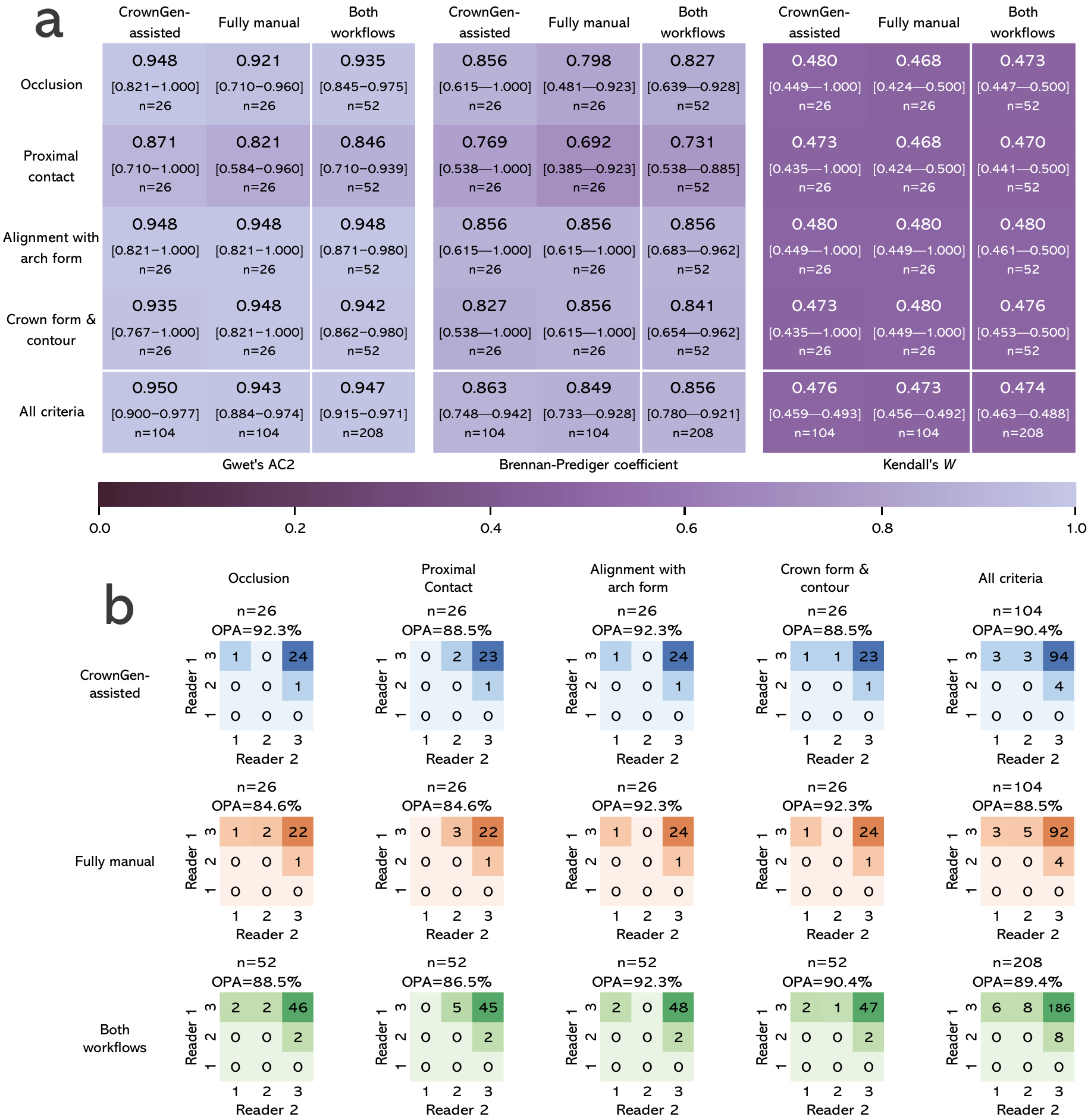}
\captionsetup{width=0.9\textwidth, skip=7pt}
\caption{\textbf{Inter-reader agreement.} Agreements are presented at four aggregation levels: stratified by workflow and criterion, pooled by criterion, pooled by workflow, and overall. \textbf{a.} Agreement metrics, including Gwet’s AC2, Brennan–Prediger coefficient, and Kendall’s $W$. 95\% CIs are derived from nonparametric bootstrap ($n=10,000$ iterations). \textbf{b.} Contingency matrices cross-tabulating the two readers’ ordinal scores. Diagonal cells indicate exact agreement. Values are raw counts, with the overall percent agreement (OPA; $\sum_i n_{ii}/\sum_{i,j} n_{ij}\times100\%$) reported for each matrix.}\label{fig_irr}
\end{figure}

\subsubsection{Inter-rater Reliability}
To ensure the robustness of our quality evaluation, we assessed the consistency between the two readers. The primary metric, Gwet’s AC2~\cite{gwet,gwet2014definitive}, yielded a coefficient of 0.947 when computed across all individual ratings, indicating excellent agreement. High agreement was consistently observed across all subsets of criteria and workflows, with AC2 values ranging from 0.821 to 0.948 (Figure~\ref{fig_irr}a). Sensitivity analyses using Kendall's $W$ (0.474), the Brennan-Prediger coefficient (0.856), and overall percentage agreement (89.4\%; Figure~\ref{fig_irr}b) further support a moderate to substantial level of inter-rater reliability.

\section{Discussion}\label{sec3}
In this work, we introduced CrownGen, a generative framework that automates patient-customized crown design using a diffusion model on a novel tooth-level representation. We demonstrated that CrownGen generates multiple, anatomically harmonized prostheses in a single inference pass, irrespective of their number, type, or location. This capability directly addresses the primary bottleneck in digital dentistry: the time-consuming, artisanal process of manual crown design~\cite{xie2025morphological, hosseinimanesh2025personalized}. By providing a fully formed initial design respecting the patient's unique occlusal environment, CrownGen can drastically reduce design turnaround, lower costs, and improve patient access to care.

This work makes several key contributions to generative AI for digital prosthodontics. First, CrownGen is, to our knowledge, the first AI framework capable of generating a variable number of crowns in a single pass. This contrasts starkly with existing methods architecturally constrained to single-tooth generation~\cite{hwang2018learning,tian2021dcpr,tian2022dual,yuan2020personalized,chau2024accuracy,ding2023morphology,farook2023computer,hosseinimanesh2025personalized,yang2024dcrownformer,lessard2022dental} and point cloud completion models whose performance collapses in multi-tooth restoration scenarios~\cite{pointsea,adapointr,proxyformer}. By overcoming this limitation, CrownGen provides a generalized solution for a wide spectrum of clinical needs. We evaluated restorations of up to six missing teeth, a threshold that captures the vast majority of fixed-restoration cases~\cite{alenezi2023technical, campbell2017removable}. In practice, once the edentulous span exceeds 4$\sim$5 consecutive teeth, the intervention paradigm shifts to removable partial dentures~\cite{alenezi2023technical, campbell2017removable} whose design priorities (major connectors, rests, clasps, base extension) differ fundamentally from the anatomic crown modelling of fixed restorations addressed by CrownGen~\cite{campbell2017removable}. Thus, our evaluation aligns with prevailing clinical practice within the fixed-prosthodontic domain and tests the model across a full spectrum of relevant clinical challenges.

Second, CrownGen introduces a paradigm shift in representing the dentition. Rather than treating the dental arch as a monolithic geometric input, we decompose it into a constellation of individual tooth objects. This tooth-centric approach enables our DITA mechanism to explicitly model inter-tooth relationships through a spatially-weighted attention system~\cite{shaw2018self, harvey2022flexible}, learning to prioritize morphogenetic signals from adjacent and opposing teeth (Supplementary Figure~\ref{fig_dita}) for superior anatomical accuracy. This discrete representation is the key innovation unlocking multi-tooth generation, as the model can dynamically differentiate context from target teeth for generation. It also provides a crucial clinical advantage: defective teeth near a restoration site can be selectively excluded from the input context to prevent them from negatively influencing the outcome. Existing arch-level models lack this mechanism for targeted intervention, making CrownGen a more robust and clinically intelligent system.

Third, CrownGen offers unparalleled scalability in data utilization. Existing AI models typically require curated training pairs of a context scan and a single, human-designed crown~\cite{hwang2018learning,tian2021dcpr,tian2022dual,yuan2020personalized,ding2023morphology,farook2023computer,hosseinimanesh2025personalized,yang2024dcrownformer,lessard2022dental}. The acquisition of such specific data is slow and resource-intensive. While some methods leverage natural teeth as ground-truth~\cite{chau2024accuracy}, all are hampered by the need for complete, \enquote{perfect} dental scans for training data, as missing or defective teeth in the input would compromise the learning process, especially if such an anomaly appears near the crown to be generated. This requirement for pristine data leads to the discarding of most real-world clinical scans because they are typically acquired from patients who visit the clinic due to their oral problems. CrownGen’s architecture and our self-bootstrapping training strategy circumvent this. By using an initial model to generate plausible \enquote{pseudo-crowns} for edentulous sites in our training data, we effectively recover incomplete scans, transforming a large, suboptimal dataset into a vast training resource. The ability to leverage imperfect clinical data at scale is a profound practical advantage, ensuring CrownGen's continuous improvement and adaptation.

The practical implementation of CrownGen is guided by several deliberate choices that merit discussion. First, while diffusion models are known for their computational intensity during inference~\cite{song2020denoising}, CrownGen’s generative module requires approximately 85 seconds per pass on an NVIDIA RTX 4090 GPU. Although slower than non-diffusion models, this is a fraction of manual design time, and critically, the inference time remains constant regardless of the number of crowns generated, making the overhead negligible in complex multi-tooth restoration cases. Second, CrownGen's workflow presupposes an initial tooth segmentation step. We view this as a pragmatic design choice rather than a barrier. AI-driven tooth segmentation is a mature technology already integrated into widely adopted dental CAD software like ExoCAD and Medit, which provide clinically-validated segmentation with user-friendly interfaces for minor corrections, minimizing additional overhead. Finally, CrownGen is designed to generate the supragingival anatomy of the crown, as it operates on a collection of teeth already segmented from intraoral scans. This approach remains agnostic to the underlying restorative interface, granting its broad applicability. Consequently, a final adaptation step is required to morph the cervical portion of the generated design so that it conforms precisely to the restoration’s margin line. For the scope of our work, this is a deliberate design choice, as the geometry of margin lines is extraordinarily diverse and dependent on highly variable factors, such as the unpredictable geometry of a prepared tooth or the specific dimensions and shapes of a manufactured implant abutment~\cite{goodacre2001tooth, minye2018preparation}. Attempting to model this vast combinatorial space may not be the most effective approach. Instead, CrownGen integrates seamlessly with existing dental CAD platforms (e.g., ExoCAD, CEREC, 3Shape Dental System). These platforms have robust functions that can delineate the margin line from the jaw with minimal user intervention and automatically morph the initial crown design to adapt to the margin line. CrownGen provides the anatomically harmonized initial proposal—a creatively demanding and time-intensive part of the process—which can then be finalized using established clinical tools. This synergistic workflow leverages the strengths of both AI generation and proven CAD tools, offering a powerful, immediately applicable solution for accelerating digital dental workflows.

In conclusion, CrownGen represents a significant step towards the complete automation of prosthetic dental design. By moving beyond the rigid, single-object completion paradigm, our framework provides a flexible, scalable, and clinically robust solution capable of addressing complex, multi-tooth restorative cases.

\begin{figure}[H]
\centering
\includegraphics[width=1.0\textwidth]{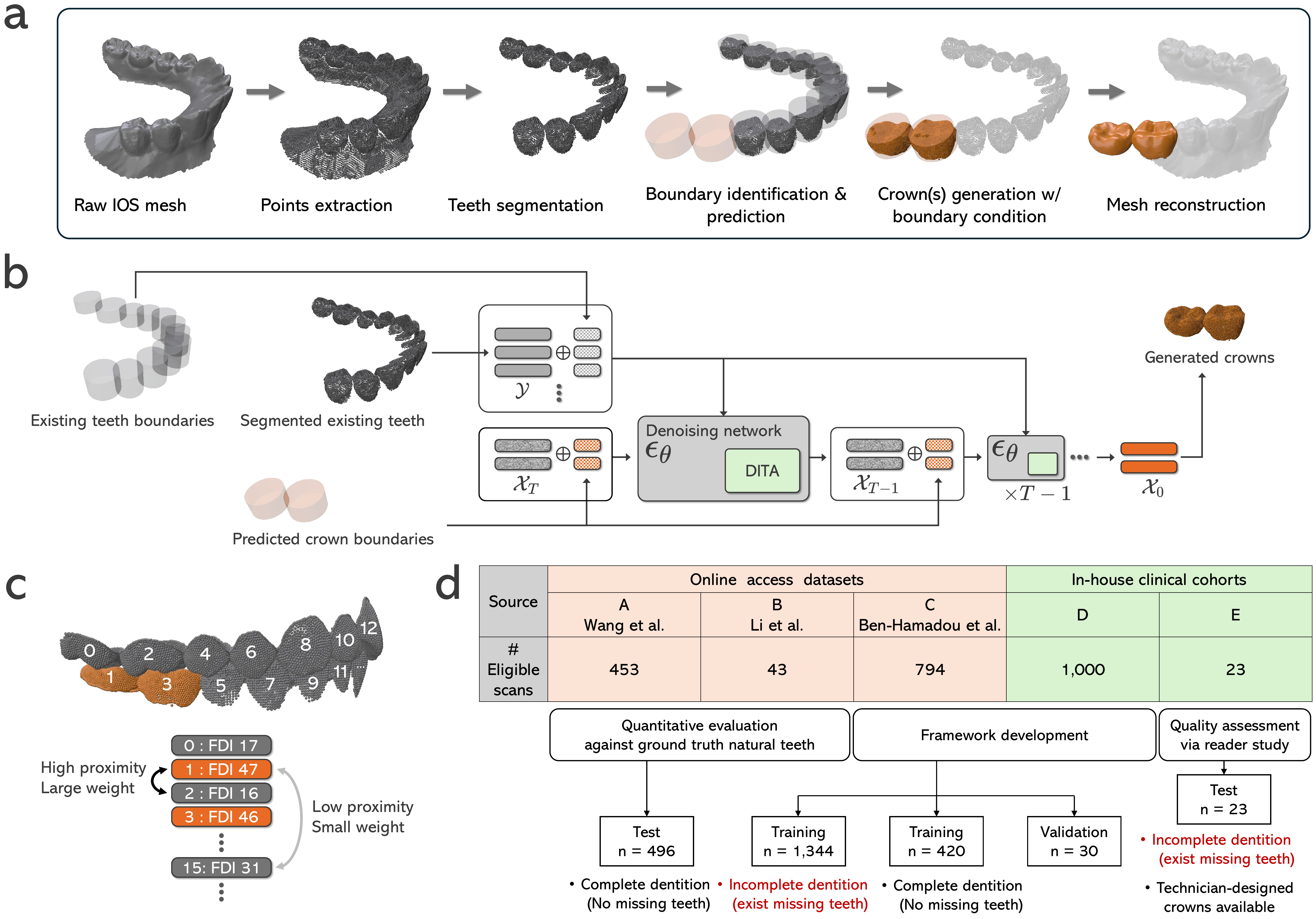}
\captionsetup{width=1.0\textwidth, skip=7pt} 
\caption{\textbf{Overview of CrownGen.} \textbf{a.} End-to-end inference pipeline of CrownGen. \textbf{b.} Architecture of the core generative module. The denoising network $\epsilon_\theta$ processes the complete set of teeth, which includes context teeth $\mathcal{Y}$ and target crowns $\mathcal{X}$ initialized as Gaussian noise. Each tooth's point cloud is concatenated with its corresponding cylindrical boundary. The network's DITA layers explicitly model morphogenetic relationships by learning attention weights informed by the relative distances between teeth. \textbf{c.} Illustration of the sequential tooth indexing method employed by the DITA layers to organize and process the constellation of teeth within the dental arch. \textbf{d.} Composition of our multi-source dataset. The development of CrownGen utilized 1,794 scans, comprising 430 fully dentate and 1,364 partially edentulous scans. Quantitative evaluation was conducted on a curated set of 496 healthy dentition scans, while the clinical reader study involved data from 23 patients that included technician-designed prostheses.}\label{fig_overall}
\end{figure}

\section{Methodology}\label{sec4}

\subsection{Data Acquisition and Preparation}\label{sec4sub1}
The development and evaluation of CrownGen were performed using a large-scale, multi-source dataset comprising 2,310 retrospective intraoral scans (Figure~\ref{fig_overall}d, Supplementary Table~\ref{tab_demographic}). This cohort was aggregated from three public online repositories (Sources A~\cite{wang20243d}, B~\cite{li2024fine}, and C~\cite{toothfairy1, toothfairy2}) and two private clinical cohorts (Sources D and E). Each intraoral scan consisted of digital 3D mesh files representing the patient's upper and lower jaws. The private clinical data was sourced from the Delun Dental Hospital Group, a large multi-center network of over 30 independent branches throughout the Greater Bay Area of China. The development cohort, Source D, contains scans collected between 2022 and 2024 from several of these clinical sites. Our clinical reader study utilized a separate cohort (Source E) comprising 26 restoration cases from 23 anonymized patients who underwent dental restoration treatment, collected from a dedicated branch in 2025. This cohort was unique in that it contained both the prepared, pre-treatment dentition and the final technician-designed crown(s) for each case, making it the ideal basis for our reader study.

A standardized pre-processing pipeline was applied to all data to ensure consistency. A key step was ensuring complete segmentation of the individual tooth and occlusal alignment. While Sources A, B, D, and E provided scans with pre-occluded jaws, Source C consisted of separated, non-occluded jaws. We recovered the optimal bite registration for these scans using an algorithm that maximizes surface contact between the jaws, guided by the clinical Midline-Canine-Molar (MCM) relationship~\cite{deng2020automatic, chang2010automatic}. This process required dental landmarks, which were partially available in Source C~\cite{toothfairy1, toothfairy2} and were manually annotated for the remaining scans by trained dentists. Furthermore, as scans from the clinical cohorts (Sources D and E) lacked segmentation labels, we employed a standalone segmentation network based on the Point Transformer~\cite{pointtransformer} architecture, which was trained on the expert-segmented scans provided in Source C. To ensure ground-truth quality, all inferred annotations were subsequently inspected by a trained dentist, and any suboptimal segmentation labels were manually rectified using Meshmixer (Meshmixer 3.5; Autodesk, San Francisco, USA). Finally, all fully processed dentitions were manually positioned to a standardized coordinate system using Blender (Blender 4.1; Blender Foundation, Amsterdam, Netherlands), where the center of the four maxillary incisors was positioned at the origin, the facial direction was oriented along the negative $Y$ axis, the sagittal plane was aligned with the $YZ$ plane, and the occlusal plane rested on the $XY$ plane.

The pre-processed data was partitioned for distinct purposes. The primary development set for training and validating CrownGen and its learnable mesh reconstruction module~\cite{dpsr} was compiled from Sources C and D. To ensure the anatomical fidelity, this development pool underwent a rigorous two-stage curation overseen by trained orthodontists. First, scans were excluded if they presented with six or more missing or severely anomalous teeth, with the count based on a 28-tooth permanent dentition excluding third molars. Second, the remaining scans were assessed for functional integrity. Scans exhibiting severe malocclusion (e.g., extensive crossbite, open bite), or excessive crowding and spacing that compromised the natural arch form were also excluded. For cases with minor to moderate malalignment deemed correctable by orthodontic standards, we employed a validated simulation model to digitally reposition the teeth into an ideal arch form and occlusal relationship~\cite{wei2020tanet}. This curation resulted in a final development cohort of 1,794 scans, comprising 430 fully dentate scans and 1,364 scans with at least one missing tooth, reflecting a clinically realistic data distribution. 10 fully dentate scans and 20 partially edentulous scans were randomly selected and held out for validation.

For quantitative evaluation, we established an external test set using Sources A and B to benchmark generated crowns against ground-truth natural teeth. To ensure that our benchmarks were based on an anatomically ideal standard, this evaluation set was meticulously curated to contain healthy dentitions. Trained orthodontists manually inspected each scan and included only complete dentitions (all 28 teeth present) exhibiting healthy morphology, proper alignment, and ideal occlusal relationships. Source A, comprising post-orthodontic treatment scans, was particularly valuable for this purpose~\cite{wang20243d}. This rigorous filtering yielded a final evaluation set of 496 scans. This curated set provides an unbiased measure of generative quality in the context of clinical workflow, since any anomalous context teeth can be selectively omitted from CrownGen's input. Finally, the clinical cohort (Source E) was reserved exclusively for the reader study.

\subsection{Related Work in AI-driven Crown Generation}\label{sec4sub2}
To contextualize the contributions of this work, we provide a detailed review of prior art. Early explorations in this domain translated three-dimensional intraoral scans into two-dimensional occlusal-depth images, reframing crown design as an image synthesis task~\cite{hwang2018learning, tian2021dcpr, tian2022dual}. Pioneering work by Hwang et al. applied a generative adversarial network (GAN) to predict crown-filled depth scans, learning from both human-designed prostheses and the spatial profiles of opposing teeth~\cite{hwang2018learning}. Building on this, Tian et al. proposed a two-stage conditional GAN that first restored the primary occlusal relationship and then enriched surface details using a specialized groove-parsing network~\cite{tian2021dcpr}. However, these 2D methods are inherently constrained, failing to capture the full volumetric complexity of a tooth and leaving a substantial portion of the design workload unaddressed.

To overcome these limitations, subsequent research shifted to three-dimensional representations~\cite{chau2024accuracy, ding2023morphology, farook2023computer}. Voxel-based approaches employed 3D GANs to synthesize crowns as occupancy grids. For instance, Ding et al. utilized a 3D deep convolutional GAN to generate personalized crowns~\cite{ding2023morphology}, while Chau et al. similarly applied a 3D GAN to reconstruct missing molars from incomplete dental casts~\cite{chau2024accuracy}. Despite representing a conceptual advance, voxelization suffers from the curse of dimensionality, becoming computationally intractable at the high resolutions needed for clinical fidelity and sacrificing fine anatomical details.

More recent studies have converged on point cloud representations. Accordingly, most contemporary crown generation models function as specialized point cloud completion networks~\cite{hosseinimanesh2025personalized, yang2024dcrownformer, lessard2022dental}. Hosseinimanesh et al. proposed a model that encodes the local geometry of the input context into feature vectors, which are then processed by a transformer to predict the missing crown~\cite{hosseinimanesh2025personalized}. Yang et al. introduced DCrownFormer, a point-to-mesh transformer that generates a crown mesh directly from the input point cloud, using a morphology-aware cross-attention module and a curvature-penalty loss to enhance occlusal detail~\cite{yang2024dcrownformer}.

\subsection{CrownGen: Generative Module}\label{sec4sub3}
Denoising diffusion probabilistic models (DDPMs) are a class of generative models that synthesize data by learning to reverse a fixed, finite‐length Markov chain of corruption. A forward (diffusion) process progressively perturbs a clean data sample $\mathcal X_0\!\sim\!q(\mathcal X_0)$ into nearly isotropic Gaussian noise $\mathcal X_T\!\approx\!\mathcal N(\mathbf 0,\mathbf I)$ over $T$ timesteps. A neural network is then trained to parameterize the reverse (denoising) process, which maps the noise back to the original data distribution.

Each fully dentate training scan is decomposed into two non-overlapping sets of point cloud teeth: (i) a set of context teeth $\mathcal Y$ and (ii) a set of $1\le k \le 6$ target crowns $\mathcal X=\{x^{(1)},\dots,x^{(k)}\}$ to be generated. To enable the model to differentiate between these roles, we assign a binary indicator (0 for context, 1 for target) concatenated as a feature to every point in each tooth point cloud. The corresponding cylindrical bounds $\mathcal{B}$ for each target crown are provided as an additional spatial condition. Finally, to provide the model with explicit information on anatomical identity and position, the 3D coordinates of each point are augmented with an 8-dimensional embedding of the tooth's unique FDI identifier.

For the set of target crowns $\mathcal X$, we define the forward diffusion chain as:
\begin{equation}
q(\mathcal X_{1:T}\mid\mathcal X_0)
\;=\;\prod_{t=1}^{T}
q(\mathcal X_t\mid\mathcal X_{t-1}),
\label{eq:forward-markov}
\end{equation}
with each transition defined by a linear–Gaussian step:
\begin{equation}
q(\mathcal X_t\mid\mathcal X_{t-1})
=\mathcal N\!\bigl(\sqrt{\alpha_t}\,\mathcal X_{t-1},
                   \beta_t\mathbf I\bigr), \qquad 
\alpha_t = 1-\beta_t.
\label{eq:forward-step}
\end{equation}
We use a linear variance schedule for the noise parameters $\{\beta_t\}_{t=1}^T$:
\begin{equation}
\beta_t
=\beta_{\text{min}}
+\bigl(\beta_{\text{max}}-\beta_{\text{min}}\bigr)\,
\frac{t-1}{T-1},
\qquad
\beta_{\text{min}}=10^{-4},\;
\beta_{\text{max}}=2\times10^{-2},\;
T=1000.
\end{equation}
Because \eqref{eq:forward-step} is Gaussian, we simply sample $\mathcal X_t$ directly from $\mathcal X_0$:
\begin{equation}
q(\mathcal X_t\mid\mathcal X_0)
=
\mathcal N\!\bigl(\sqrt{\bar\alpha_t}\,\mathcal X_0,
                  (1-\bar\alpha_t)\mathbf I\bigr),
\qquad
\bar\alpha_t:=\prod_{s=1}^{t}\alpha_s.
\label{eq:closed-form-forward}
\end{equation}
Since the time‑reversal of \eqref{eq:forward-markov} is intractable, we approximate it with a learnable reverse process:
\begin{equation}
p_\theta(\mathcal X_{0:T}\mid\mathcal Y,\mathcal{B})
=
p(\mathcal X_T)\;
\prod_{t=1}^{T}
p_\theta(\mathcal X_{t-1}\mid\mathcal X_t,\mathcal Y,\mathcal{B}),
\label{eq:reverse-joint}
\end{equation}
where the prior $p(\mathcal X_T)=\mathcal N(\mathbf 0,\mathbf I)$ and each transition is parameterized as:
\begin{equation}
p_\theta(\mathcal X_{t-1}\mid\mathcal X_t,\mathcal Y,\mathcal{B})
=
\mathcal N\!\bigl(\boldsymbol\mu_\theta(\mathcal X_t,\mathcal Y,\mathcal{B},t),
                  \tilde\beta_t\mathbf I\bigr).
\label{eq:reverse-step}
\end{equation}
The variance is fixed to its optimal value for Gaussian forward processes:
\begin{equation}
\tilde\beta_t
=
\frac{1-\bar\alpha_{t-1}}{1-\bar\alpha_t}\,\beta_t.
\end{equation}

\noindent Following Ho et al.~\cite{diffusion1ho}, we then parameterize the mean $\boldsymbol\mu_\theta$ by training a denoising network $\boldsymbol\epsilon_\theta$ to predict the noise component $\boldsymbol\epsilon\sim\mathcal N(\mathbf 0,\mathbf I)$ from the noisy sample $\mathcal X_t$:
\begin{equation}
\boldsymbol\mu_\theta(\mathcal X_t, \mathcal Y, \mathcal B, t) =
\frac{1}{\sqrt{\alpha_t}}
\Bigl(
\mathcal X_t
-\frac{\beta_t}
        {\sqrt{1-\bar\alpha_t}} \boldsymbol\epsilon_\theta(\mathcal X_t,\mathcal Y,\mathcal{B},t)
\Bigr).
\label{eq:mu-from-eps}
\end{equation}
To train the denoising network, we optimize the variational lower bound (VLB) on the negative log-likelihood. The standard VLB objective can be decomposed into a sum of Kullback–Leibler (KL) divergence terms:
\begin{align}
\mathcal L_{\text{VLB}} 
&= \mathbb E_{q} \bigl[ \underbrace{D_{KL}\bigl( q(\mathcal{X}_T|\mathcal{X}_0) \,\|\, p(\mathcal{X}_T) \bigr)}_{L_T} + \sum_{t=2}^{T} \underbrace{D_{KL}\bigl( q(\mathcal{X}_{t-1}|\mathcal{X}_t, \mathcal{X}_0) \,\|\, p_{\theta}(\mathcal{X}_{t-1}|\mathcal{X}_t, \mathcal Y, \mathcal B) \bigr)}_{L_{t-1}} \underbrace{- \log p_{\theta}(\mathcal{X}_0|\mathcal{X}_1, \mathcal Y, \mathcal B) }_{L_0} \bigr].
\label{eq:vlb}
\end{align}
Ho et al. showed that each KL term in this objective can be re-weighted and simplified, leading to a much simpler surrogate objective that yields higher-quality samples~\cite{diffusion1ho}. By ignoring the weighting terms, this objective reduces to a straightforward mean-squared error between the true and predicted Gaussian noise:
\begin{equation}
\mathcal L
=
\mathbb E_{\substack{\mathcal X_0\sim q(\mathcal X_0)\\[2pt]
                     t\sim\mathcal U\{1..T\}\\[2pt]
                     \boldsymbol\epsilon\sim\mathcal N(\mathbf 0,\mathbf I)}}
\Bigl[
\bigl\|
\boldsymbol\epsilon
-
\boldsymbol\epsilon_\theta(
  \sqrt{\bar\alpha_t}\,\mathcal X_0 + \sqrt{1-\bar\alpha_t}\,\boldsymbol\epsilon,
  \mathcal Y,\mathcal{B},t)
\bigr\|_2^2
\Bigr].
\label{eq:mse-loss}
\end{equation}
During training, the loss is computed only on the point clouds of the target crowns $\mathcal{X}$, using a binary mask to ignore the context teeth $\mathcal{Y}$, which remain noise-free and serve only as a condition. To generate new crowns, we start by sampling from the prior, $\mathcal X_T \sim \mathcal N(\mathbf 0,\mathbf I)$, within the predicted boundaries $\mathcal B$. We then iteratively denoise this sample for $t=T, \dots, 1$ using the learned reverse process:
\begin{equation}
\mathcal X_{t-1}
=
\frac{1}{\sqrt{\alpha_t}}
\Bigl(
\mathcal X_t
- \frac{\beta_t}
      {\sqrt{1-\bar\alpha_t}} \boldsymbol\epsilon_\theta(\mathcal X_t,\mathcal Y,\mathcal{B},t)
\Bigr)
+\sqrt{\tilde\beta_t}\,\boldsymbol\eta,
\qquad
\boldsymbol\eta\sim
\begin{cases}
\mathcal N(\mathbf 0,\mathbf I), & t>1,\\[4pt]
\mathbf 0, & t=1.
\end{cases}
\end{equation}
This iterative denoising procedure yields the final set of generated crowns, $\mathcal{X}_0$.

Our denoising network, $\boldsymbol\epsilon_\theta$, is built upon a PointNet++ encoder–decoder backbone~\cite{pointnet++} in which each PointNet substructure is replaced with point–voxel convolution (PVC) operators~\cite{pvc, pvd}. The architecture is U-shaped and symmetric, comprising four set abstraction (SA) blocks that sequentially downsample each input tooth cloud to a set of latent centroids, followed by four feature propagation (FP) blocks that hierarchically upsample and fuse multi-resolution features. Because each element in the input set $\mathcal X\cup\mathcal Y$ is an individual tooth-level point cloud, we process each tooth independently through the SA–FP pipeline by treating them as separate samples in the batch dimension. The resulting latent descriptors are then aggregated for inter-tooth processing. Full architectural details of the denoising network are provided in Supplementary Figure~\ref{fig_ddpm}.

The crucial inter-tooth relationships are learned via our proposed Distance-weighted Inter-Tooth Attention (DITA) layers, which are inserted after each SA and FP block and at the architectural bottleneck. DITA augments the standard multi-head self-attention mechanism~\cite{multiheadattention} with learnable, index-based relative positional encodings (RPEs)~\cite{shaw2018self, harvey2022flexible}.

Each DITA layer operates on an input latent tensor:
\begin{equation}
\mathbf Z^{\mathrm{in}}\!=\!\bigl[\mathbf z^{\mathrm{in}}_1,\dots,\mathbf z^{\mathrm{in}}_K\bigr]\in\mathbb R^{K\times C}
\end{equation}
where each row vector $\mathbf z_i^{\mathrm{in}}$ is the feature descriptor for a single tooth and $K=|\mathcal{X}|+|\mathcal{Y}|$. 
The indices of the teeth are arranged in a fixed, linearized zig-zag FDI sequence (Figure~\ref{fig_overall}c): ${17,\,47,\,16,\,46,\dots,11,\,41,\,21,\,31,\dots,27,\,37}$. This ordering of the maxillary and mandibular arches ensures that the index difference, $\Delta_{ij}=i-j$, serves as a simple yet effective proxy for anatomical distance. Specifically, we adopt the RPE implementation from~\cite{harvey2022flexible}, where for every tooth pair $(i,j)$, we compute this discrete distance and transform it into a 3-dimensional vector representation $\mathbf r_{ij}$:
\begin{equation}
\mathbf r_{ij}=
\bigl[\log(1+\max(\Delta_{ij},0)),\;
      \log(1+\max(-\Delta_{ij},0)),\;
      \mathbf{1}_{\Delta_{ij}=0}\bigr]\in\mathbb R^{3},
\end{equation}
where the positive and negative distances are treated symmetrically, and self-relations ($\Delta_{ij}=0$) are explicitly flagged. A multi-layer perceptron then projects $\mathbf r_{ij}$ into three learnable RPE bias vectors: $\{\mathbf p^{\mathrm Q}_{ij},\mathbf p^{\mathrm K}_{ij},\mathbf p^{\mathrm V}_{ij}\}$. Queries ($\mathbf q_i$), keys ($\mathbf k_i$), and values ($\mathbf v_i$) are obtained via linear projections of the per-tooth feature vector $\mathbf z^{\mathrm{in}}_i$. Within each attention head, DITA computes the attention scores $e_{ij}$ and weights $\alpha_{ij}$ as follows:
\begin{equation}
e_{ij} \;=\; 
\frac{1}{\sqrt F}\,\mathbf q_i^{\!\top}\mathbf k_j
\;+\;
\mathbf q_i^{\!\top}\mathbf p^{\mathrm K}_{ij},
\;+\;
\mathbf p^{\mathrm Q\!\top}_{ij}\mathbf k_j
\qquad
\alpha_{ij}=\frac{\exp(e_{ij})}{\sum_{k=1}^{K}\exp(e_{ik})},
\end{equation}
where $F$ is the feature dimension of the keys. The output feature vector for each tooth is then computed as a residual connection on the weighted sum of values:
\begin{equation}
\mathbf z^{\mathrm{out}}_i
=
\mathbf z^{\mathrm{in}}_i
+
\sum_{j=1}^{K}\alpha_{ij}\bigl(\mathbf v_j+\mathbf p^{\mathrm V}_{ij}\bigr).
\end{equation}

For each training iteration, we simulate diverse clinical scenarios by randomly masking 1 to 6 teeth from a complete 28-tooth permanent dentition. These masked teeth are assigned to the target set $\mathcal{X}$, with the remainder assigned to the context set $\mathcal{Y}$. Each tooth in $\mathcal{X} \cup \mathcal{Y}$ is represented by a point cloud of 1024 points, sampled uniformly from its mesh surface. To enhance model generalization, we employ a suite of data augmentations: (i) random shuffling of points within each tooth's point cloud; (ii) bilateral mirroring of the entire dentition with a corresponding remapping of FDI indices; and (iii) isotropic scaling of the dentition within the range $[0.95, 1.05]$. A dropout rate of 0.1 is applied within each PVC operator.

Training proceeds in two stages. In the first stage, the generative module is trained for 3000 epochs on a dataset of 420 fully dentate scans using the Adam optimizer~\cite{adam} with an initial learning rate of $4\times10^{-5}$, which is decayed by a factor of 0.4 at epoch 1500. In the second stage, we use the pre-trained model from the first stage to generate \enquote{pseudo-crowns} for the partially edentulous clinical scans, thereby creating a large, fully dentate, and anatomically consistent augmented dataset. The model is then fine-tuned on this combined dataset for an additional 2400 epochs with an initial learning rate of $2\times10^{-5}$ and a stepwise decay of 0.45 every 800 epochs.

The generated point clouds are converted into clinically usable manifold meshes using a learnable surface reconstruction model based on Differentiable Poisson Surface Reconstruction (DPSR)~\cite{dpsr}. Given a generated point cloud, this network first predicts an upsampled and oriented point cloud with per-point normal vectors. These oriented points are then input to a differentiable Poisson solver that computes a 3D indicator function on a grid, specifying the interior and exterior of the shape. The final, watertight mesh is extracted from this grid using the Marching Cubes algorithm. The reconstruction network is trained independently on ground-truth tooth meshes by optimizing a mean squared error loss between the predicted and ground-truth indicator grids.

\subsection{CrownGen: Boundary Prediction Module}\label{sec4sub4}
The generative module requires a set of cylindrical bounds $\mathcal{B}$ to spatially constrain and localize the synthesis of each target crown. These bounds are provided by an independent, preceding boundary prediction module. This lightweight deep learning model is trained to analyze the context dentition $\mathcal{Y}$ and regress the optimal cylindrical volume for each missing tooth.

This task is analogous to 3D object detection~\cite{pan20213d, yang2018pixor}, but instead of conventional bounding boxes, we use cylindrical bounds, which better capture the natural morphology of teeth. A key prerequisite is our data pre-processing step, which orients each dentition's occlusal plane parallel to the $XY$ plane, thereby aligning the primary longitudinal axis of each tooth with the $Z$ axis. Under this standardized alignment, the ground-truth bound for a tooth is derived by: (i) projecting its points onto the $XY$ plane; (ii) fitting a minimal enclosing circle to these points to obtain the radius $r$ and center $(c_x, c_y)$; and (iii) using the minimum and maximum $Z$ coordinates of the tooth to define its height $h = z_{\text{max}}-z_{\text{min}}$ and axial center $c_z = (z_{\text{min}}+z_{\text{max}})/2$. Because the cylinder is rotationally invariant in the occlusal view, this parameterization is compactly defined by just five scalars—center coordinates $(c_x, c_y, c_z)$, radius $r$, and height $h$—providing a computationally efficient and geometrically intuitive prior for crown generation.

The architecture of the boundary prediction module mirrors the encoder of our main denoising network, comprising three SA blocks with PVC operators and DITA layers to effectively model inter-tooth relationships. The training data is prepared similarly to that for the generative module: for each complete dentition, 1 to 6 teeth are randomly selected as targets. Each tooth is represented by a cloud of 512 points; for the target teeth, these points are zeroed out to maintain a consistent input tensor shape while signaling their absence. The DITA layers are thereby constrained to compute attention scores only over the non-zero feature vectors of the context teeth $\mathcal{Y}$. The model is supervised on the pre-computed ground-truth cylindrical parameters of the masked teeth.

The model's objective is to regress a parameter tensor $\mathcal{B}_{\text{pred}} \in \mathbb{R}^{|\mathcal{X}| \times 5}$, containing the five predicted cylinder parameters for each of the $|\mathcal{X}|$ target teeth. The loss is computed only on these target parameters using a binary mask. We employ the Smooth $L1$ loss, which is less sensitive to outliers than the standard $L2$ loss:
\begin{equation}
\mathcal{L}_{\text{bound}} = \frac{1}{|\mathcal{X}|} \sum_{i \in \mathcal{X}} \text{smooth}_{L1}(\mathcal{B}_{\text{pred}, i} - \mathcal{B}_{\text{gt}, i}),
\end{equation}
where $\mathcal{B}_{\text{pred}, i}$ and $\mathcal{B}_{\text{gt}, i}$ are the predicted and ground-truth parameter vectors for the $i$-th target tooth, respectively, and the loss function is defined as:
\begin{equation}
\text{smooth}_{L1}(x) =
\begin{cases}
0.5 x^2, & \text{if } |x| < 1, \\
|x| - 0.5, & \text{otherwise}.
\end{cases}
\end{equation}

The same data augmentation techniques used for the generative module are employed here. The boundary prediction module is trained for 1000 epochs using the Adam optimizer~\cite{adam} with a dropout rate of 0.3. The learning rate was initialized to $3 \times 10^{-4}$ and decayed to $3 \times 10^{-6}$ over the course of training using a cosine annealing schedule.

\subsection{Quantitative Benchmarking Protocol}

\subsubsection{Comparative State-of-the-art Methods}
A direct comparison with existing AI-driven crown generation methods is confounded by a fundamental architectural mismatch. State-of-the-art crown generation models are extensions of the conventional point cloud completion paradigm~\cite{hosseinimanesh2025personalized, yang2024dcrownformer, lessard2022dental}. They are architecturally designed to accept a single, non-separable point cloud dentition data as input and generate a single crown, typically anchored to a prepared abutment tooth. Consequently, they are fundamentally ill-equipped to handle the generalized clinical problem of generating a variable number of crowns from a variable-cardinality input of context teeth.

To establish a rigorous and fair comparison, we therefore selected three state-of-the-art general-purpose point cloud completion networks from the computer vision domain: PointSea~\cite{pointsea}, AdaPoinTr~\cite{adapointr}, and ProxyFormer~\cite{proxyformer}. PointSea utilizes multi-view depth projections to recover global structure and fine details. AdaPoinTr employs a Transformer architecture~\cite{multiheadattention} with adaptive query generation and a geometry-aware denoising objective. ProxyFormer intelligently partitions the input into \enquote{existing} and \enquote{missing} proxy points that interact via a specialized Transformer to refine the prediction.

To adapt these single-object completion models to our multi-tooth generation task, we trained six separate instances of each comparing method—one for each possible number of target missing teeth to generate, from one to six. This approach provided each method with a scenario-specific advantage, as each model was optimized for a fixed-cardinality task. In stark contrast, a single, unified CrownGen model was tasked with handling all scenarios without modification or retraining. To ensure a level playing field, all comparing methods were trained on the exact same expanded pseudo-crown dataset as CrownGen.

\subsubsection{Quantitative Evaluation of Geometric Fidelity}
The quantitative evaluation rigorously benchmarks the geometric fidelity of generated crowns against their ground-truth natural tooth counterparts. The assessment proceeds in two phases: first at the level of the raw point cloud, and second on the reconstructed mesh surfaces. For the point cloud evaluation, we employed three standard metrics: Chamfer Distance $L1$ (CD), Earth Mover's Distance (EMD), and the $F1$ score. 
CD and EMD quantify the dissimilarity between two point clouds. CD computes the symmetrized mean of absolute nearest-neighbor distances between the sets. EMD computes the minimum optimal-transport \enquote{work} to transform one point set into the other, capturing global geometric correspondence. The $F1$ score provides a nuanced assessment of reconstruction quality by combining precision (the fraction of generated points close to the ground-truth) and recall (the fraction of ground-truth points captured by the generation)~\cite{tatarchenko2019single}. We evaluate $F1$ at distance thresholds of 0.3 mm, 0.5 mm, and 1.0 mm, reflecting clinically relevant tolerances for prosthetic fit.

While CrownGen operates on a point cloud for computational efficiency, the final clinical product is a mesh. Therefore, we conducted a second evaluation phase on the reconstructed mesh surfaces. We used Average Surface Distance (ASD) to measure the mean Euclidean distance between the generated and ground-truth mesh surfaces, providing a direct assessment of overall geometric accuracy and fit. Complementing this, Normal Consistency (NC) evaluates the congruence of local surface orientation by computing the cosine similarity between the surface normal vectors of corresponding points

\subsubsection{Test Scenario Design}
Each test scenario was created by randomly omitting a variable number of teeth from the fully dentate dentition to serve as the generation targets. Our evaluation strategy employed two distinct suites of test scenarios to assess performance at both the raw point cloud and final reconstructed mesh levels.
The first suite was created for point cloud-level evaluation, comprising a total of 26,288 test scenarios (10, 14, 10, 9, 5, and 5 scenarios for restoring 1 to 6 missing teeth, respectively). A key methodological consideration for this suite was ensuring a fair benchmark against baseline models, which can only generate a single, non-delineated point cloud that is difficult to separate without creating ambiguous or fused geometry (Figure~\ref{fig_pointcloud_all}b). Therefore, for analyses on specific functional groups in restoration scenarios with more than one missing tooth (i.e., all missing teeth were anteriors, premolars, or molars), we utilized a dedicated subset of these scenarios (at least 9, 6, 6, 3, and 3 scenarios for restoring 2 to 6 missing teeth, respectively) where teeth were randomly selected for omission from within a single functional group. Evaluation for this entire suite was conducted on a per-scenario basis, with the unified ground-truth point cloud of all missing teeth treated as a single sample.

To complement this benchmark, we designed a second suite specifically for evaluating the mesh-level performance of CrownGen and its ablated variants. This evaluation, which is unique to our framework due to its ability to generate distinct crowns, simulated complex and clinically realistic restoration needs by selecting the missing teeth for generation at random from anywhere in the dental arch. This protocol consisted of 6, 3, 2, 1, 1, and 1 unique scenarios per dentition involving restorations of one to six missing teeth, respectively, yielding 6,944 test scenarios. Our sampling ensured that every one of the 28 teeth in a given dentition was selected for omission at least once across the scenarios generated from it. As our framework generates distinct prostheses, this suite enabled a more granular per-tooth evaluation, comprising a total of 16,368 individual restored crowns across all scenarios.

\subsection{Clinical Reader Study}\label{sec4sub5}

\subsubsection{Study Cohort and Design Workflows}
The reader study was designed as a direct, workflow-centric comparison of crown design quality and efficiency. The study utilized a diverse, retrospective cohort of 23 real-world patients with 26 distinct crown restoration cases. (Supplementary Table~\ref{tab_reader_demographic}). For each case, two crown mesh variants were generated via two independent CAD workflows (Figure~\ref{fig_reader_workflow}).        

For the fully manual CAD workflow, crowns were designed by three highly experienced prosthodontic laboratory technicians. Two of the technicians had over a decade of physical wax-up experience, complemented by 5 and 4 years of digital CAD work, respectively. The third technician was a digital design specialist with 5 years of focused CAD experience. The patients were randomly distributed among them (8, 8, and 7 cases, respectively), and each crown was designed using a standard tooth template library. For the CrownGen-assisted CAD workflow, a single general dental technician with entry-level proficiency (0.5 years) performed only the post-processing of the initial designs generated by CrownGen to yield the final crowns (Supplementary Figure~\ref{fig_postprocess}). For both workflows, all design and finalization tasks were executed using ExoCAD (DentalCAD 3.2 Elefsina; EXOCAD GmbH, Darmstadt, Germany).

Post-processing steps for both workflows included margin line delineation; automatic margin adaptation that morphs the cervical portion of the initial design to the margin line; and final surface refinement and smoothing. Active design time for the fully manual workflow was measured from the initial import of the patient's scan to the completion of the final design. Active design time for the CrownGen-assisted workflow included the computational inference for both the segmentation model and CrownGen, in addition to the manual time required for post-processing.

\subsubsection{Crown Quality Assessment}
Clinical assessment was performed by two residency-trained dentists with 4 and 5 years of clinical experience, respectively. To mitigate memory and comparison biases, we adopted a cross-over evaluation protocol involving two reading sessions separated by a 14-day washout period. For each restoration, the design from one workflow was randomly assigned to be evaluated in the first session, and the corresponding design from the other workflow was evaluated in the second session~\cite{park2023methods, anderson2024deep}. This ensured that in the first session, half the viewed restorations were from the fully manual workflow and half were from the CrownGen-assisted workflow. To minimize sequence-related biases, the assignment of design to each session and the presentation order of restorations within each session were independently randomized for each of the two readers. All finalized crown designs were presented to the readers within the context of the full dental arch using 3D Slicer version 5.6.2 under standardized viewing conditions on a color-calibrated medical-grade monitor.

\subsubsection{Workload-Calibrated Digital Crown Quality Rubric}
To quantify design quality in a manner directly relevant to clinical efficiency, we developed a workload-calibrated digital crown quality rubric in consultation with experienced prosthodontists (Supplementary Table~\ref{tab_rubric}). The rubric's framework is designed to stratify the quality of digital crown designs based on the anticipated downstream manual workload required to render them clinically viable for manufacturing. While established systems~\cite{cvar2005reprint,hickel2023revised} guide evaluation of post-placement, in-vivo prostheses, our study targets pre-manufacturing digital designs. Thus, we retain their anatomical and functional constructs but recast the anchors to reflect design readiness and edit burden in the digital environment. Designs were rated on a 3-point Likert scale, where each score corresponds to a distinct level of required clinical intervention:

\begin{itemize}
    \item \textbf{Score 3: Clinically acceptable, minimal to no adjustment.} The design is anatomically and functionally sound, meeting all critical clinical requirements. It is considered ready for manufacturing, requiring at most minor surface refinements that do not alter the core morphology, thereby representing a near-complete automation of the design task.

    \item \textbf{Score 2: Clinically correctable, targeted morphological adjustment.} The design possesses a valid foundation but exhibits moderate deficiencies in form or function. It necessitates targeted morphological adjustments by a skilled technician to become clinically acceptable. While serving as a viable starting point, it requires a non-trivial investment of manual labor to rectify its shortcomings.

    \item \textbf{Score 1: Clinically unacceptable, complete redesign} The design contains fundamental flaws that violate core anatomical or functional principles. Correction would require a complete redesign from inception, rendering the initial output clinically unusable. This score signifies a failure to produce a design that reduces the manual workload.

\end{itemize}

\subsubsection{Non-inferiority Analysis}
A non-inferiority (NI) test was conducted to assess whether crowns produced from the CrownGen-assisted CAD workflow were at least as good as those produced from a fully manual, expert-led CAD workflow. For each case, the reader-averaged score for the CrownGen-assisted workflow was subtracted from the reader-averaged score for the fully manual workflow (CrownGen-assisted$-$Fully manual), yielding a paired difference of scores per case. The primary estimand was the mean of these paired differences across the 26 cases.

Before analysis, we prespecified NI margin on the 1$-$3 reader score using a 5-percentage-point absolute margin (5 pp) over the full scale, corresponding to $0.10$ points ($5\% \times(3-1)=0.10$). The CrownGen-assisted workflow was defined as non-inferior if the lower bound of the 95\% CI for the mean difference in scores was greater than $-0.10$. The 5 pp margin was chosen a priori as a conservative threshold, informed by dental restoration trials that commonly permit 10$-$15 pp differences for crown quality endpoints~\cite{olegario2022stainless,sterzenbach2025randomised,verma2024evaluation}. Since the prior evidence and endpoints from related prosthodontic studies are not directly transferable to our AI-versus-manual, reader-rated setting, we adopted a conservative margin tailored to this context and justified on clinical and statistical grounds~\cite{EMA, FDA, mauri2017challenges, ICH}. Accordingly, 5 pp was selected to reflect a tighter standard for functional crown quality than is customary in related literature. NI was concluded when the two-sided 95\% lower confidence bound for the mean paired difference exceeded $-0.10$ points.

To ensure robustness, two different methods were used to compute the 95\% CI: (1) a nonparametric, case-level bootstrap CI ($10,000$ iterations), prespecified as the primary analysis for determining NI; and (2) a standard $t$-based CI, reported as a sensitivity analysis. The bootstrap method, which avoids normality assumptions, was included to account for the discrete, ordinal nature of the reader scores and the high prevalence of tied scores, which can affect the assumptions of the $t$-distribution. The $t$-based CI is presented to assess robustness to distributional assumptions without altering the primary decision rule.

\subsubsection{Statistical Design and Reliability Considerations}
Because no prior data were available, the sample size ($n=26$ paired cases) was determined by feasibility. Primary inference relied on the NI criterion—namely, that the lower bounds of the two-sided 95\% confidence intervals for the mean paired difference exceeded the prespecified margin of $0.10$ points on the 1–3 composite scale, thereby demonstrating NI across all endpoints. We additionally performed a retrospective assurance (precision) analysis based on the observed standard deviations of the paired differences. For the composite endpoint, the observed standard deviation was $0.06$, implying an assurance of 80\% power for a minimum detectable NI margin of $0.034$ points. For the occlusion endpoint, which had the largest standard deviation ($0.13$), the analogous threshold for the NI margin was $0.074$ points—still comfortably below the prespecified margin of $0.10$.

To assess inter-rater reliability, we used raw agreement counts, Kendall’s coefficient of concordance (Kendall's $W$), the Brennan–Prediger coefficient, and Gwet’s AC2 (Figure~\ref{fig_irr}). Kendall’s $W$ is rank-based and can be conservative when ties are frequent—as in our data, where scores largely cluster at 3. Brennan–Prediger coefficient is a chance-corrected agreement that assumes equal category prevalence~\cite{brennan}, mitigating prevalence bias relative to Kappa and thus well-suited to our imbalanced, tie-heavy paired ratings. We designate Gwet’s AC2 as the primary metric because it adjusts for chance under class imbalance and performs well with ordinal categories~\cite{gwet,gwet2014definitive}. We did not estimate the intraclass correlation coefficient because the study included fewer than three raters and fewer than 30 cases~\cite{icc_guideline}; nor did we report Cohen’s Kappa owing to its well-known “Kappa paradox,” in which high prevalence of a single category can depress Kappa despite substantial agreement~\cite{gwet}.

\subsection{Implementation Details}\label{sec4sub6}
All analyses in this study were performed using Python (version 3.11.0). The deep learning components of the CrownGen framework, including the diffusion-based generative module and the boundary prediction network, were implemented using PyTorch (version 2.5.1). Core data manipulation and scientific computing were conducted using NumPy (version 1.26.4) and pandas (version 2.2.3). Statistical analyses were performed with SciPy (version 1.12.0). Three-dimensional point cloud and mesh visualizations were generated using Polyscope (version 2.2.1), while two-dimensional plots and figures were created with Matplotlib (version 3.8.3) and Seaborn (version 0.13.2).

\section*{Data Availability}
This study utilized both private in-house clinical cohorts and publicly available online access datasets for system development and evaluation. The private data, collected and managed under institutional supervision, are accessible under specific conditions. Interested researchers may request access for non-commercial academic use by contacting the corresponding author. All requests are subject to institutional review and require a documented material transfer agreement. The public datasets consisted of intraoral scans previously released by Wang et al.~\cite{wang20243d}, Li et al.~\cite{li2024fine}, and Ben Hamadou et al.~\cite{toothfairy1, toothfairy2}.

\section*{Ethics Approval}
This study received ethical approval from the Human and Artefacts Research Ethics Committee (HAREC), as documented in protocol HREP-2024-0257.

\section*{Acknowledgements}
This work was supported by the Hong Kong Innovation and Technology Commission (Project No. GHP/006/22GD and ITCPD/17-9), and the Research Grants Council of the Hong Kong Special Administrative Region, China (Project No. T45-401/22-N).

\bibliography{sn-bibliography}

\newpage
\begin{appendices}
\section*{Supplementary Information}

\vspace*{\fill}

\begin{figure}[H]
\centering
\includegraphics[width=1.0\textwidth]{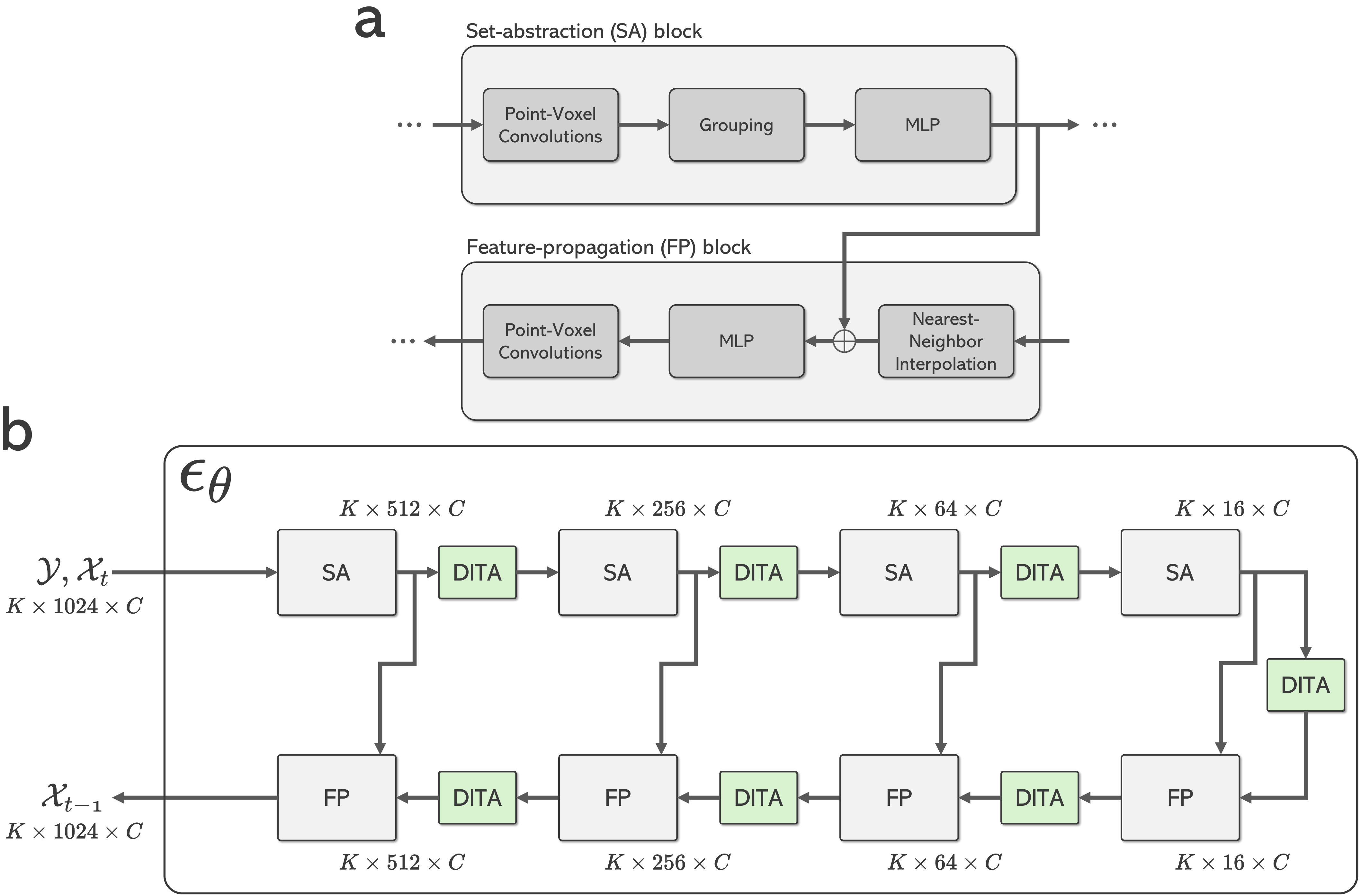}
\captionsetup{width=1.0\textwidth, skip=7pt} 
\caption{\textbf{Architecture of the denoising network of CrownGen.} \textbf{a.} The fundamental building blocks of the denoising network. Set abstraction (SA) modules hierarchically down-sample the point cloud while encoding local geometric features. Feature propagation (FP) modules up-sample features via interpolation. Skip connections propagate high-resolution features from the encoder (SA blocks) to the decoder (FP blocks) ~\cite{pointnet++}. \textbf{b.} The CrownGen denoising network employs a U-Net-like encoder-decoder architecture with four levels of SA and FP blocks. Distance-weighted Inter-Tooth Attention (DITA) layers are interleaved between the blocks. These DITA layers enable the network to explicitly model the spatial relationships between all teeth at multiple feature scales, allowing it to condition the generation of each crown on its precise anatomical context.}\label{fig_ddpm}
\end{figure}

\vspace*{\fill}
\newpage

\vspace*{\fill}

\begin{figure}[H]
\centering
\includegraphics[width=1.0\textwidth]{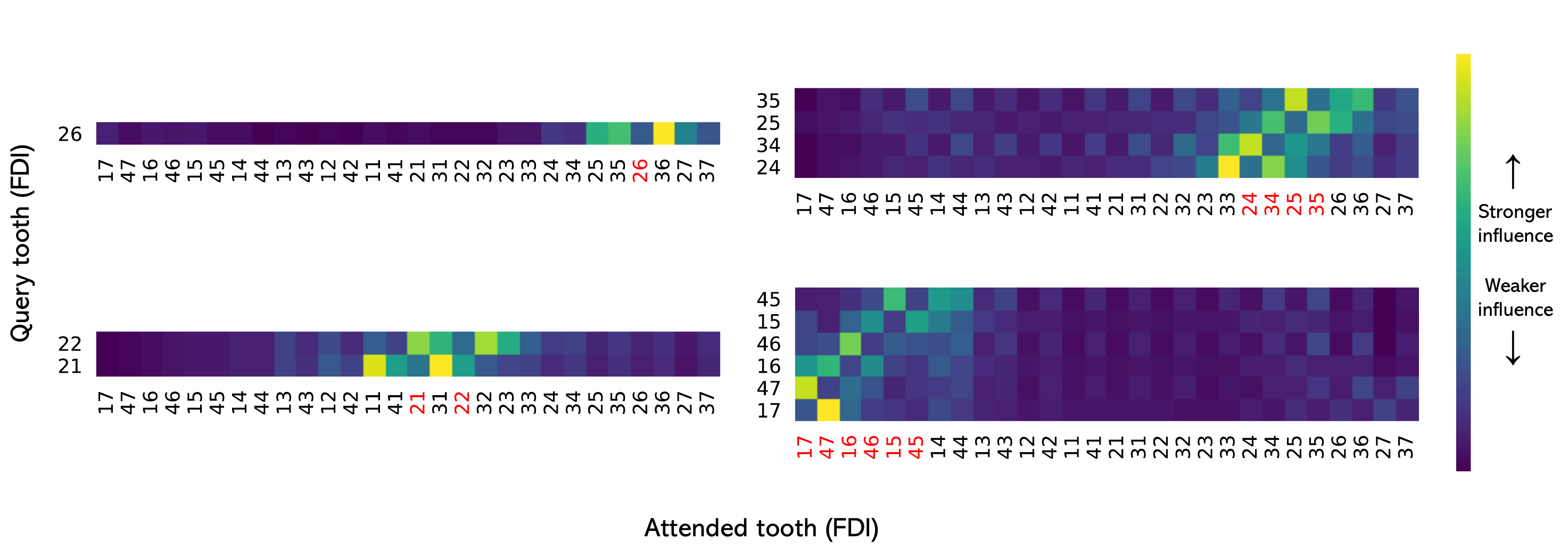}
\captionsetup{width=1.0\textwidth, skip=7pt} 
\caption{\textbf{Visualization of the Distance-weighted Inter-tooth Attention (DITA) mechanism.} Attention heatmaps for four distinct restoration scenarios, illustrating the influence of existing context teeth on the crowns being generated. Each row represents a query from a target tooth position, while columns represent the context teeth being attended to. Attention weights were captured during early-stage denoising (timesteps $t=[1000,800]$) and averaged across all DITA layers, attention heads, and batches. Color intensity corresponds to the mean attention weight (brighter = stronger influence). The heatmaps confirm that DITA learns clinically meaningful relationships, concentrating its attention on immediate mesial/distal neighbors and opposing antagonistic teeth, which are critical for determining correct occlusal and proximal morphology.}\label{fig_dita}
\end{figure}

\vspace*{\fill}
\newpage

\begin{figure}[H]
\centering
\includegraphics[width=1.0\textwidth]{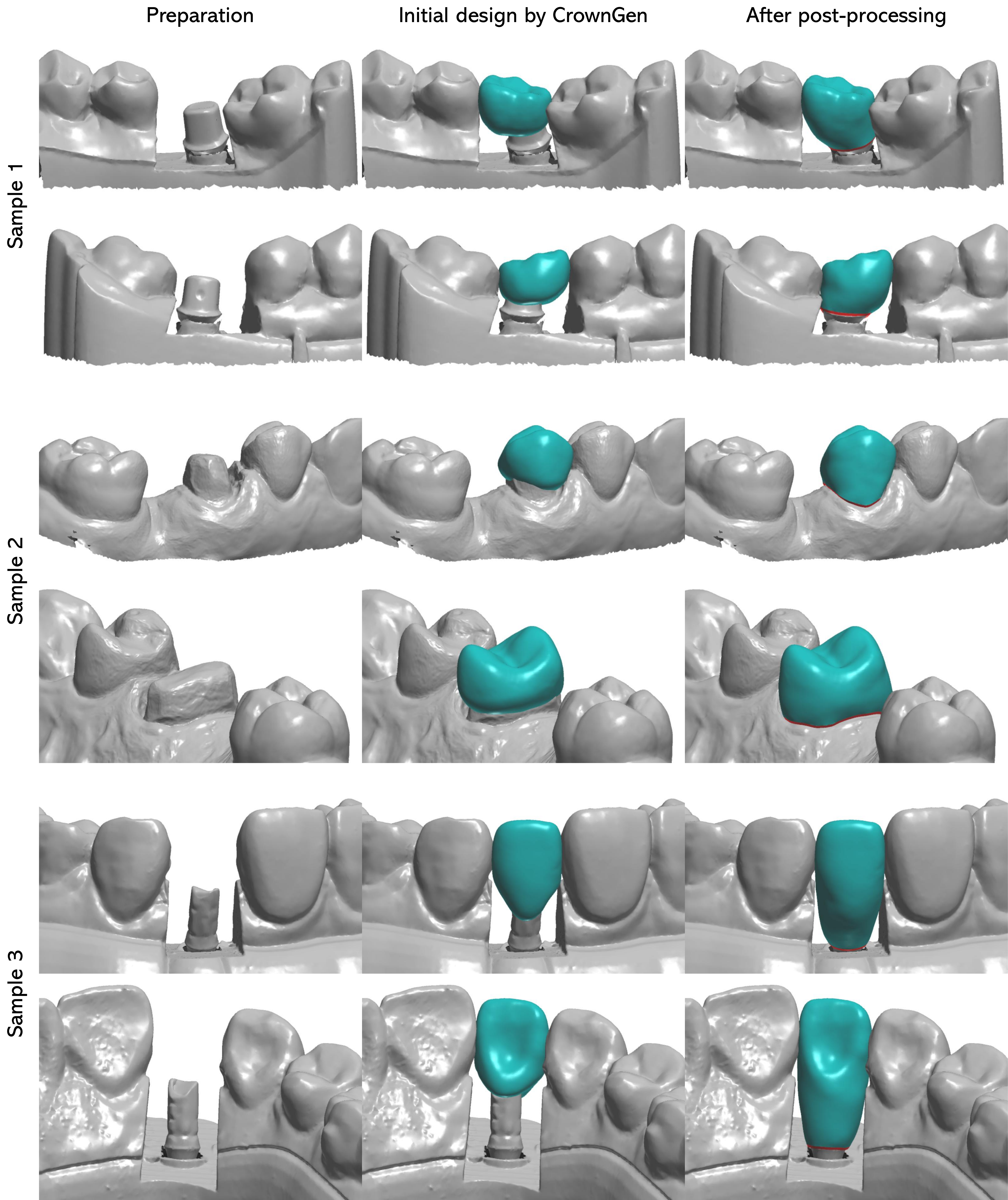}
\captionsetup{width=1.0\textwidth, skip=7pt} 
\caption{\textbf{Finalization of CrownGen's initial designs.} Visual examples demonstrating the minimal post-processing required to finalize CrownGen's output. \textbf{Left:} The patient's prepared jaw. \textbf{Center:} The initial crown design generated by CrownGen, which already possesses the complete, patient-customized supragingival morphology. \textbf{Right:} The finalized crown after post-processing. The finalization step primarily involves margin line detection and the semi-automated adaptation of the crown's cervical portion to this margin. These steps are routine elements of the digital workflow, fully supported by standard tools in commercial dental CAD systems. By automating the complex anatomical design phase, CrownGen reduces the finalization to a straightforward, technician-guided task, significantly lowering the demand for specialized, from-scratch design expertise. Visualizations were created using the 3Shape 3D Viewer (version 1.3).}\label{fig_postprocess}
\end{figure}

\begin{figure}[H]
  \centering
\includegraphics[width=0.82\textwidth]{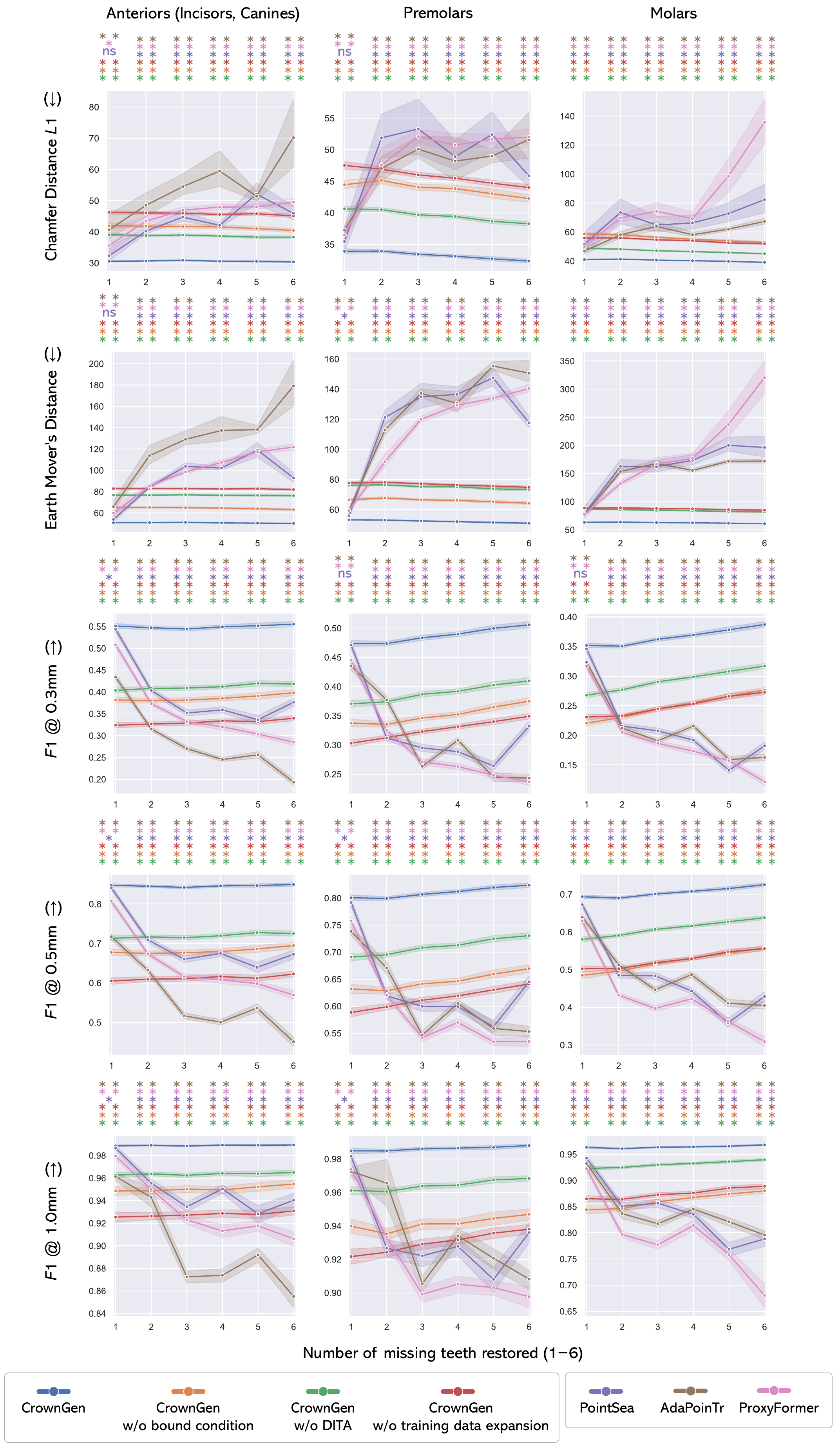}
\end{figure}
\begin{figure}[H]
    \caption{\textbf{Quantitative evaluation of generated point cloud crowns by functional group.} Mean performance (curves) with 95\% CIs (shaded bands; nonparametric bootstrap, $10,000$ resamples) of CrownGen, its ablated variants, and state-of-the-art methods on a dedicated subset of scenarios derived from 496 test dentitions. In these scenarios, all missing teeth were constrained to a single functional group (anterior, premolar, or molar). Performance is stratified by group and the number of missing teeth. Metrics marked with ↓ indicate lower is better, and vice versa. Chamfer distance and Earth mover’s distance are scaled by $10^{3}$. Statistical differences relative to CrownGen were assessed with a two-sided paired $t$-test ($^\ast p < 0.05$, $^{\ast\ast} p < 0.01$). Detailed sample sizes for each condition are provided in Supplementary~\cref{tab_pointcloud_anterior,tab_pointcloud_premolar,tab_pointcloud_molar}.}\label{fig_pointcloud_toothgroup}
\end{figure}

\begin{sidewaystable}[h]
\renewcommand{\arraystretch}{1.5}
\begin{tabular*}{\textwidth}{@{\extracolsep\fill}llllll}
\toprule%
& \multicolumn{3}{c}{\textbf{Online access datasets}} & \multicolumn{2}{c}{\textbf{In-house clinical cohorts}} \\
\cmidrule(lr){2-4} \cmidrule(lr){5-6}
\textbf{Characteristic} & \textbf{Source A} & \textbf{Source B} & \textbf{Source C} & \textbf{Source D} & \textbf{Source E} \\
\midrule
Type & Public~\cite{wang20243d} & Public~\cite{li2024fine} & Public~\cite{toothfairy1, toothfairy2} & Private & Private \\
\addlinespace
Role &\Centerstack[l]{External test set\\(Geometric fidelity)}&\Centerstack[l]{External test set\\(Geometric fidelity)}&\Centerstack[l]{Model training\\\& validation}&\Centerstack[l]{Model training\\\& validation}&\Centerstack[l]{Clinical evaluation\\(Reader study)}\\
\addlinespace
Scans available &1060&90&900&1000&23\\
\addlinespace
\Centerstack[l]{Scans included\\(post-curation)}   &453&43&794&1000&23\\
\addlinespace
Age distribution                   

&
8-35 years: 100\%
&
\Centerstack[l]{
0-10 years: 4\%\\
11-20 years: 27\%\\
21-30 years: 48\%\\\
31-40 years: 18\%\\
41-50 years: 3\%}
&
\Centerstack[l]{
$<$16 years: 70\%\\
16-59 years: 27\%\\
$>$60 years: 3\%}
&
\Centerstack[l]{
11-20 years: 0.3\%\\
21-30 years: 9\%\\
31-40 years: 17.7\%\\
41-50 years: 22\%\\
51-60 years: 28\%\\
61-70 years: 17.6\%\\
$>$70 years: 5.4\%}
&
\Centerstack[l]{Unavailable\\(Anonymized)}\\
\addlinespace

Sex distribution &Not specified&Not specified&\Centerstack[l]{Male: 50\%\\Female: 50\%}&\Centerstack[l]{Male: 51.5\%\\Female: 47.7\%\\Unknown: 0.8\%}&\Centerstack[l]{Unavailable\\(Anonymized)}\\
\addlinespace
Acquisition device(s) &\Centerstack[l]{iTero\\(Align Technology)}&\Centerstack[l]{TRIOS 3\\(3Shape)}&\Centerstack[l]{Primescan\\(Dentsply Sirona)\\TRIOS 3\\(3Shape)\\iTero\\(Align Technology)}&\Centerstack[l]{Aoralscan 3\\(Shining 3D)}&\Centerstack[l]{TRIOS 3\\(3Shape)}\\
\addlinespace
Geographic origin           &China&China&France, Belgium&China&China\\
\addlinespace
Segmentation masks &\Centerstack[l]{Expert-annotated\\(Provided)}&\Centerstack[l]{Expert-annotated\\(Provided)}&\Centerstack[l]{Expert-annotated\\(Provided)}&\Centerstack[l]{Model-generated\\(Expert-verified)}&\Centerstack[l]{Model-generated\\(Expert-verified)}\\
\addlinespace
Data collection period            &2017$-$2023&Not specified&Not specified&2022$-$2024&2025\\

\botrule
\end{tabular*}
\captionsetup{skip=7pt}
 \caption{\textbf{Overview of data sources for model development and evaluation.} Summary of public and in-house clinical data cohorts used in this study. For public datasets, demographic information pertains to the original, unfiltered cohorts as per-scan data was unavailable; the exact distribution of the curated subsets may differ. For in-house clinical cohorts (Sources D and E), data were provided after initial pre-screening by the collaborating clinic; therefore, all available scans met the inclusion criteria.}\label{tab_demographic}
\end{sidewaystable} 

\begin{table}[ht]
    \centering
    \footnotesize
    \makebox[\textwidth][c]{%
  \begin{minipage}{\textwidth}
    \renewcommand{\arraystretch}{1.2}
      \begin{tabular}{@{}c@{}llll ll @{}}
        \toprule
            \multicolumn{1}{c}{\Centerstack{Number of\\missing teeth\\restored}} 
          &  \multicolumn{1}{c}{} 
          & \multicolumn{1}{c}{\Centerstack{Chamfer\\Distance $L1$ (↓)}} 
          & \multicolumn{1}{c}{\Centerstack{Earth Mover's\\Distance (↓)}} 
          & \multicolumn{1}{c}{$F1$ $@$ 0.3 mm (↑)}
          & \multicolumn{1}{c}{$F1$ $@$ 0.5 mm (↑)}
          & \multicolumn{1}{c}{$F1$ $@$ 1.0 mm (↑)} \\
        \midrule
        \multirow{7}{*}{\shortstack{1\\($n = 4,960$)}}
          & CrownGen            & \textbf{34.532} $\mathbf{(34.309\text{-}34.766)}$ & \textbf{55.150} $\mathbf{(54.816\text{-}55.499)}$ & \textbf{0.471} $\mathbf{(.468\text{-}.475)}$ & \textbf{0.790} $\mathbf{(.786\text{-}.793)}$ & \textbf{0.980} $\mathbf{(.979\text{-}.981)}$\\
          & \tiny w/o bound cond. & 47.485 $(46.957\text{-}48.047)$ & 72.027 $(71.302\text{-}72.787)$ & 0.323 $(.319\text{-}.327)$ & 0.609 $(.604\text{-}.615)$ & 0.916 $(.913\text{-}.920)$\\
          & \tiny w/o DITA            & 42.314 $(42.022\text{-}42.610)$ & 79.522 $(78.991\text{-}80.058)$ & 0.355 $(.351\text{-}.359)$ & 0.669 $(.664\text{-}.673)$ & 0.951 $(.949\text{-}.953)$\\
          & \tiny w/o data expansion  & 49.397 $(48.998\text{-}49.799)$ & 83.062 $(82.508\text{-}83.624)$ & 0.291 $(.288\text{-}.295)$ & 0.571 $(.566\text{-}.576)$ & 0.907 $(.904\text{-}.910)$\\
          & PointSea            & 38.782 $(36.550\text{-}41.456)$ & 61.777 $(58.513\text{-}65.758)$ & 0.466 $(.461\text{-}.471)$ & 0.778 $(.774\text{-}.783)$ & 0.973 $(.970\text{-}.975)$\\
          & AdaPoinTr           & 41.439 $(39.927\text{-}43.877)$ & 67.471 $(65.835\text{-}69.981)$ & 0.403 $(.398\text{-}.407)$ & 0.701 $(.696\text{-}.706)$ & 0.956 $(.954\text{-}.959)$\\
          & ProxyFormer         & 39.980 $(38.038\text{-}42.805)$ & 64.774 $(62.608\text{-}67.828)$ & 0.435 $(.430\text{-}.439)$ & 0.742 $(.737\text{-}.746)$ & 0.962 $(.959\text{-}.964)$\\
        \midrule
        \multirow{7}{*}{\shortstack{2\\($n = 6,944$)}}
          & CrownGen            & \textbf{35.111} $\mathbf{(34.953\text{-}35.266)}$ & \textbf{55.813} $\mathbf{(55.577\text{-}56.047)}$ & \textbf{0.462} $\mathbf{(.459\text{-}.464)}$ & \textbf{0.781} $\mathbf{(.779\text{-}.784)}$ & \textbf{0.979} $\mathbf{(.978\text{-}.979)}$\\
          & \tiny w/o bound cond. & 48.085 $(47.755\text{-}48.419)$ & 73.119 $(72.652\text{-}73.593)$ & 0.318 $(.315\text{-}.321)$ & 0.603 $(.600\text{-}.607)$ & 0.912 $(.910\text{-}.914)$\\
          & \tiny w/o DITA            & 42.429 $(42.240\text{-}42.613)$ & 79.723 $(79.375\text{-}80.062)$ & 0.355 $(.352\text{-}.357)$ & 0.669 $(.666\text{-}.671)$ & 0.950 $(.949\text{-}.951)$\\
          & \tiny w/o data expansion  & 49.650 $(49.391\text{-}49.901)$ & 83.577 $(83.206\text{-}83.938)$ & 0.291 $(.289\text{-}.293)$ & 0.570 $(.567\text{-}.573)$ & 0.905 $(.903\text{-}.907)$\\
          & PointSea            & 53.854 $(51.444\text{-}56.520)$ & 116.852 $(113.708\text{-}120.239)$ & 0.318 $(.315\text{-}.321)$ & 0.613 $(.609\text{-}.617)$ & 0.916 $(.913\text{-}.918)$\\
          & AdaPoinTr           & 49.009 $(47.726\text{-}50.276)$ & 105.543 $(103.216\text{-}107.883)$ & 0.266 $(.264\text{-}.269)$ & 0.614 $(.609\text{-}.618)$ & 0.915 $(.909\text{-}.921)$\\
          & ProxyFormer         & 52.825 $(51.795\text{-}53.962)$ & 100.705 $(98.958\text{-}102.687)$ & 0.308 $(.305\text{-}.310)$ & 0.582 $(.578\text{-}.585)$ & 0.899 $(.896\text{-}.901)$\\
        \midrule
        \multirow{7}{*}{\shortstack{3\\($n = 4,960$)}}
          & CrownGen            & \textbf{34.924} $\mathbf{(34.766\text{-}35.084)}$ & \textbf{55.485} $\mathbf{(55.243\text{-}55.733)}$ & \textbf{0.465} $\mathbf{(.462\text{-}.467)}$ & \textbf{0.784} $\mathbf{(.782\text{-}.786)}$ & \textbf{0.979} $\mathbf{(.979\text{-}.980)}$\\
          & \tiny w/o bound cond. & 47.309 $(46.991\text{-}47.634)$ & 72.175 $(71.715\text{-}72.646)$ & 0.324 $(.321\text{-}.327)$ & 0.612 $(.608\text{-}.615)$ & 0.918 $(.916\text{-}.920)$\\
          & \tiny w/o DITA            & 42.018 $(41.827\text{-}42.211)$ & 79.262 $(78.896\text{-}79.631)$ & 0.361 $(.358\text{-}.363)$ & 0.675 $(.672\text{-}.678)$ & 0.952 $(.951\text{-}.953)$\\
          & \tiny w/o data expansion  & 49.058 $(48.812\text{-}49.304)$ & 82.983 $(82.616\text{-}83.354)$ & 0.297 $(.295\text{-}.299)$ & 0.577 $(.574\text{-}.580)$ & 0.909 $(.907\text{-}.911)$\\
          & PointSea            & 52.248 $(50.669\text{-}53.997)$ & 124.849 $(121.829\text{-}128.217)$ & 0.293 $(.290\text{-}.296)$ & 0.591 $(.587\text{-}.596)$ & 0.911 $(.908\text{-}.914)$\\
          & AdaPoinTr           & 56.251 $(55.198\text{-}57.514)$ & 141.183 $(139.307\text{-}143.464)$ & 0.243 $(.241\text{-}.246)$ & 0.505 $(.501\text{-}.509)$ & 0.865 $(.863\text{-}.868)$\\
          & ProxyFormer         & 55.180 $(54.081\text{-}56.599)$ & 117.175 $(115.049\text{-}119.938)$ & 0.277 $(.275\text{-}.280)$ & 0.534 $(.530\text{-}.538)$ & 0.878 $(.875\text{-}.881)$\\
        \midrule
        \multirow{7}{*}{\shortstack{4\\($n = 4,464$)}}
          & CrownGen            & \textbf{34.645} $\mathbf{(34.492\text{-}34.801)}$ & \textbf{55.114} $\mathbf{(54.880\text{-}55.356)}$ & \textbf{0.470} $\mathbf{(.467\text{-}.473)}$ & \textbf{0.788} $\mathbf{(.786\text{-}.791)}$ & \textbf{0.980} $\mathbf{(.979\text{-}.981)}$\\
          & \tiny w/o bound cond. & 47.140 $(46.815\text{-}47.483)$ & 71.955 $(71.491\text{-}72.438)$ & 0.328 $(.325\text{-}.330)$ & 0.615 $(.611\text{-}.618)$ & 0.918 $(.916\text{-}.920)$\\
          & \tiny w/o DITA            & 41.617 $(41.437\text{-}41.802)$ & 78.817 $(78.456\text{-}79.184)$ & 0.366 $(.364\text{-}.369)$ & 0.681 $(.679\text{-}.684)$ & 0.954 $(.953\text{-}.955)$\\
          & \tiny w/o data expansion  & 48.598 $(48.367\text{-}48.836)$ & 82.525 $(82.171\text{-}82.890)$ & 0.303 $(.301\text{-}.305)$ & 0.584 $(.582\text{-}.587)$ & 0.911 $(.910\text{-}.913)$\\
          & PointSea            & 50.925 $(49.741\text{-}52.364)$ & 129.480 $(127.102\text{-}132.332)$ & 0.284 $(.280\text{-}.287)$ & 0.579 $(.574\text{-}.584)$ & 0.910 $(.907\text{-}.913)$\\
          & AdaPoinTr           & 55.618 $(53.777\text{-}57.878)$ & 141.933 $(138.246\text{-}146.437)$ & 0.253 $(.250\text{-}.255)$ & 0.529 $(.525\text{-}.533)$ & 0.886 $(.883\text{-}.889)$\\
          & ProxyFormer         & 53.357 $(52.416\text{-}54.481)$ & 124.460 $(122.290\text{-}126.913)$ & 0.268 $(.265\text{-}.272)$ & 0.553 $(.548\text{-}.558)$ & 0.890 $(.887\text{-}.893)$\\
        \midrule
        \multirow{7}{*}{\shortstack{5\\($n = 2,480$)}}
          & CrownGen            & \textbf{34.479} $\mathbf{(34.282\text{-}34.680)}$ & \textbf{54.853} $\mathbf{(54.549\text{-}55.163)}$ & \textbf{0.474} $\mathbf{(.470\text{-}.477)}$ & \textbf{0.791} $\mathbf{(.788\text{-}.794)}$ & \textbf{0.980} $\mathbf{(.980\text{-}.981)}$\\
          & \tiny w/o bound cond. & 46.490 $(46.115\text{-}46.866)$ & 71.230 $(70.680\text{-}71.786)$ & 0.335 $(.331\text{-}.338)$ & 0.623 $(.618\text{-}.627)$ & 0.921 $(.919\text{-}.923)$\\
          & \tiny w/o DITA            & 41.260 $(41.030\text{-}41.495)$ & 78.255 $(77.794\text{-}78.722)$ & 0.373 $(.370\text{-}.376)$ & 0.687 $(.684\text{-}.691)$ & 0.954 $(.953\text{-}.956)$\\
          & \tiny w/o data expansion  & 48.196 $(47.910\text{-}48.485)$ & 82.300 $(81.843\text{-}82.759)$ & 0.307 $(.305\text{-}.310)$ & 0.590 $(.586\text{-}.593)$ & 0.914 $(.912\text{-}.916)$\\
          & PointSea            & 56.158 $(54.532\text{-}58.062)$ & 139.520 $(136.126\text{-}143.414)$ & 0.256 $(.251\text{-}.260)$ & 0.537 $(.530\text{-}.543)$ & 0.881 $(.876\text{-}.885)$\\
          & AdaPoinTr           & 53.746 $(53.114\text{-}54.402)$ & 156.343 $(154.714\text{-}157.977)$ & 0.220 $(.217\text{-}.224)$ & 0.504 $(.499\text{-}.510)$ & 0.880 $(.876\text{-}.883)$\\
          & ProxyFormer         & 61.433 $(58.838\text{-}64.501)$ & 148.794 $(143.705\text{-}154.767)$ & 0.245 $(.242\text{-}.247)$ & 0.510 $(.505\text{-}.514)$ & 0.872 $(.868\text{-}.876)$\\
        \midrule
        \multirow{7}{*}{\shortstack{6\\($n = 2,480$)}}
          & CrownGen            & \textbf{34.338} $\mathbf{(34.155\text{-}34.522)}$ & \textbf{54.653} $\mathbf{(54.365\text{-}54.946)}$ & \textbf{0.476} $\mathbf{(.472\text{-}.479)}$ & \textbf{0.793} $\mathbf{(.790\text{-}.796)}$ & \textbf{0.981} $\mathbf{(.980\text{-}.981)}$\\
          & \tiny w/o bound cond. & 46.275 $(45.928\text{-}46.624)$ & 71.069 $(70.545\text{-}71.598)$ & 0.339 $(.336\text{-}.342)$ & 0.626 $(.622\text{-}.630)$ & 0.922 $(.919\text{-}.924)$\\
          & \tiny w/o DITA            & 41.017 $(40.810\text{-}41.229)$ & 77.904 $(77.474\text{-}78.339)$ & 0.376 $(.373\text{-}.379)$ & 0.691 $(.688\text{-}.694)$ & 0.955 $(.954\text{-}.957)$\\
          & \tiny w/o data expansion  & 47.734 $(47.475\text{-}47.997)$ & 81.873 $(81.444\text{-}82.304)$ & 0.313 $(.310\text{-}.315)$ & 0.596 $(.593\text{-}.599)$ & 0.916 $(.914\text{-}.917)$\\
          & PointSea            & 54.342 $(52.129\text{-}56.877)$ & 121.724 $(117.447\text{-}126.679)$ & 0.310 $(.305\text{-}.314)$ & 0.602 $(.595\text{-}.608)$ & 0.903 $(.898\text{-}.907)$\\
          & AdaPoinTr           & 59.901 $(57.482\text{-}62.781)$ & 168.382 $(163.625\text{-}174.049)$ & 0.198 $(.195\text{-}.201)$ & 0.475 $(.469\text{-}.480)$ & 0.866 $(.862\text{-}.870)$\\
          & ProxyFormer         & 66.812 $(63.483\text{-}70.329)$ & 160.554 $(153.816\text{-}167.567)$ & 0.239 $(.235\text{-}.243)$ & 0.513 $(.506\text{-}.519)$ & 0.862 $(.856\text{-}.869)$\\
        \bottomrule
      \end{tabular}
    \captionsetup{skip=7pt}
      \caption{\textbf{Quantitative evaluation of generated point cloud crowns. (All tooth types)} Mean performance with 95\% CIs (nonparametric bootstrap, $10,000$ resamples) comparing CrownGen with state-of-the-art baselines and ablations, stratified by number of missing teeth restored and pooled across all tooth types. Metrics marked with $\downarrow$ indicate lower-is-better and vice versa. Chamfer Distance and Earth Mover’s Distance metrics are reported $\times10^3$. Bold denotes the best method.}\label{tab_pointcloud_all}
  \end{minipage}
  }
\end{table} 

\begin{table}[ht]
    \centering
    \footnotesize
    \makebox[\textwidth][c]{%
  \begin{minipage}{\textwidth}
    \renewcommand{\arraystretch}{1.2}
      \begin{tabular}{@{}c@{}llll ll @{}}
        \toprule
            \multicolumn{1}{c}{\Centerstack{Number of\\missing teeth\\restored}} 
          &  \multicolumn{1}{c}{} 
          & \multicolumn{1}{c}{\Centerstack{Chamfer\\Distance $L1$ (↓)}} 
          & \multicolumn{1}{c}{\Centerstack{Earth Mover's\\Distance (↓)}} 
          & \multicolumn{1}{c}{$F1$ $@$ 0.3 mm (↑)}
          & \multicolumn{1}{c}{$F1$ $@$ 0.5 mm (↑)}
          & \multicolumn{1}{c}{$F1$ $@$ 1.0 mm (↑)} \\
        \midrule
        \multirow{7}{*}{\shortstack{1\\($n = 2,093$)}}
          & CrownGen            & \textbf{30.575} $\mathbf{(30.284\text{-}30.879)}$ & \textbf{50.934} $\mathbf{(50.446\text{-}51.438)}$ & \textbf{0.551} $\mathbf{(.546\text{-}.557)}$ & \textbf{0.847} $\mathbf{(.843\text{-}.852)}$ & \textbf{0.989} $\mathbf{(.987\text{-}.990)}$\\
          & \tiny w/o bound cond. & 41.967 $(41.164\text{-}42.894)$ & 65.206 $(64.115\text{-}66.434)$ & 0.382 $(.375\text{-}.389)$ & 0.678 $(.670\text{-}.686)$ & 0.949 $(.945\text{-}.953)$\\
          & \tiny w/o DITA            & 39.099 $(38.679\text{-}39.525)$ & 76.816 $(76.019\text{-}77.619)$ & 0.404 $(.398\text{-}.410)$ & 0.713 $(.706\text{-}.720)$ & 0.963 $(.960\text{-}.965)$\\
          & \tiny w/o data expansion  & 46.262 $(45.662\text{-}46.871)$ & 82.905 $(82.007\text{-}83.808)$ & 0.324 $(.318\text{-}.330)$ & 0.605 $(.597\text{-}.613)$ & 0.925 $(.921\text{-}.930)$\\
          & PointSea            & 32.355 $(30.695\text{-}35.324)$ & 53.985 $(51.298\text{-}58.762)$ & 0.544 $(.536\text{-}.551)$ & 0.841 $(.835\text{-}.848)$ & 0.987 $(.984\text{-}.989)$\\
          & AdaPoinTr           & 40.673 $(37.944\text{-}45.669)$ & 65.926 $(62.881\text{-}71.279)$ & 0.434 $(.426\text{-}.442)$ & 0.717 $(.709\text{-}.726)$ & 0.961 $(.958\text{-}.965)$\\
          & ProxyFormer         & 35.660 $(32.830\text{-}40.926)$ & 59.922 $(56.722\text{-}65.621)$ & 0.508 $(.501\text{-}.515)$ & 0.807 $(.801\text{-}.814)$ & 0.980 $(.977\text{-}.982)$\\
        \midrule
        \multirow{7}{*}{\shortstack{2\\($n = 1,865$)}}
          & CrownGen            & \textbf{30.679} $\mathbf{(30.472\text{-}30.887)}$ & \textbf{51.006} $\mathbf{(50.646\text{-}51.369)}$ & \textbf{0.547} $\mathbf{(.543\text{-}.551)}$ & \textbf{0.845} $\mathbf{(.842\text{-}.848)}$ & \textbf{0.989} $\mathbf{(.988\text{-}.990)}$\\
          & \tiny w/o bound cond. & 41.836 $(41.353\text{-}42.341)$ & 65.235 $(64.563\text{-}65.927)$ & 0.380 $(.374\text{-}.385)$ & 0.675 $(.668\text{-}.681)$ & 0.949 $(.946\text{-}.951)$\\
          & \tiny w/o DITA            & 38.825 $(38.502\text{-}39.154)$ & 76.750 $(76.102\text{-}77.398)$ & 0.408 $(.404\text{-}.413)$ & 0.717 $(.712\text{-}.722)$ & 0.964 $(.962\text{-}.965)$\\
          & \tiny w/o data expansion  & 46.082 $(45.613\text{-}46.554)$ & 83.014 $(82.292\text{-}83.739)$ & 0.327 $(.322\text{-}.332)$ & 0.609 $(.603\text{-}.616)$ & 0.926 $(.923\text{-}.929)$\\
          & PointSea            & 40.230 $(39.385\text{-}41.298)$ & 84.270 $(83.138\text{-}85.478)$ & 0.404 $(.398\text{-}.410)$ & 0.709 $(.703\text{-}.716)$ & 0.955 $(.952\text{-}.959)$\\
          & AdaPoinTr           & 48.525 $(44.856\text{-}52.477)$ & 113.752 $(105.948\text{-}123.497)$ & 0.315 $(.310\text{-}.320)$ & 0.633 $(.625\text{-}.641)$ & 0.943 $(.934\text{-}.951)$\\
          & ProxyFormer         & 43.454 $(42.288\text{-}44.926)$ & 85.443 $(83.736\text{-}87.537)$ & 0.373 $(.368\text{-}.378)$ & 0.674 $(.668\text{-}.680)$ & 0.950 $(.947\text{-}.953)$\\
        \midrule
        \multirow{7}{*}{\shortstack{3\\($n = 1,119$)}}
          & CrownGen            & \textbf{30.900} $\mathbf{(30.662\text{-}31.151)}$ & \textbf{51.236} $\mathbf{(50.810\text{-}51.684)}$ & \textbf{0.544} $\mathbf{(.540\text{-}.549)}$ & \textbf{0.842} $\mathbf{(.838\text{-}.846)}$ & \textbf{0.988} $\mathbf{(.988\text{-}.989)}$\\
          & \tiny w/o bound cond. & 41.727 $(41.164\text{-}42.349)$ & 65.018 $(64.221\text{-}65.869)$ & 0.381 $(.376\text{-}.387)$ & 0.677 $(.670\text{-}.683)$ & 0.950 $(.947\text{-}.953)$\\
          & \tiny w/o DITA            & 38.992 $(38.631\text{-}39.370)$ & 77.108 $(76.369\text{-}77.887)$ & 0.409 $(.404\text{-}.414)$ & 0.715 $(.709\text{-}.720)$ & 0.962 $(.960\text{-}.964)$\\
          & \tiny w/o data expansion  & 45.973 $(45.502\text{-}46.470)$ & 82.833 $(82.071\text{-}83.642)$ & 0.329 $(.324\text{-}.334)$ & 0.611 $(.604\text{-}.617)$ & 0.927 $(.924\text{-}.931)$\\
          & PointSea            & 44.742 $(43.407\text{-}46.565)$ & 103.717 $(101.199\text{-}107.039)$ & 0.352 $(.345\text{-}.359)$ & 0.660 $(.652\text{-}.669)$ & 0.935 $(.930\text{-}.939)$\\
          & AdaPoinTr           & 54.501 $(51.951\text{-}58.586)$ & 129.072 $(124.032\text{-}137.065)$ & 0.270 $(.265\text{-}.276)$ & 0.517 $(.509\text{-}.525)$ & 0.872 $(.867\text{-}.878)$\\
          & ProxyFormer         & 46.920 $(46.095\text{-}47.867)$ & 98.465 $(96.797\text{-}100.238)$ & 0.332 $(.327\text{-}.338)$ & 0.616 $(.609\text{-}.623)$ & 0.923 $(.919\text{-}.927)$\\
        \midrule
        \multirow{7}{*}{\shortstack{4\\($n = 1,030$)}}
          & CrownGen            & \textbf{30.600} $\mathbf{(30.389\text{-}30.811)}$ & \textbf{50.699} $\mathbf{(50.296\text{-}51.109)}$ & \textbf{0.549} $\mathbf{(.545\text{-}.554)}$ & \textbf{0.846} $\mathbf{(.843\text{-}.849)}$ & \textbf{0.989} $\mathbf{(.989\text{-}.990)}$\\
          & \tiny w/o bound cond. & 41.596 $(41.003\text{-}42.288)$ & 64.596 $(63.770\text{-}65.494)$ & 0.385 $(.380\text{-}.391)$ & 0.680 $(.673\text{-}.686)$ & 0.949 $(.946\text{-}.952)$\\
          & \tiny w/o DITA            & 38.680 $(38.342\text{-}39.021)$ & 76.584 $(75.844\text{-}77.325)$ & 0.412 $(.407\text{-}.417)$ & 0.720 $(.715\text{-}.725)$ & 0.964 $(.962\text{-}.966)$\\
          & \tiny w/o data expansion  & 45.594 $(45.128\text{-}46.064)$ & 82.560 $(81.797\text{-}83.330)$ & 0.334 $(.329\text{-}.339)$ & 0.616 $(.610\text{-}.623)$ & 0.929 $(.925\text{-}.932)$\\
          & PointSea            & 42.120 $(41.398\text{-}42.870)$ & 102.193 $(100.455\text{-}103.966)$ & 0.359 $(.352\text{-}.366)$ & 0.675 $(.667\text{-}.683)$ & 0.950 $(.946\text{-}.954)$\\
          & AdaPoinTr           & 59.494 $(54.527\text{-}65.986)$ & 137.420 $(127.414\text{-}150.373)$ & 0.245 $(.241\text{-}.250)$ & 0.501 $(.494\text{-}.507)$ & 0.874 $(.868\text{-}.880)$\\
          & ProxyFormer         & 47.977 $(47.036\text{-}48.954)$ & 107.479 $(105.137\text{-}109.869)$ & 0.320 $(.313\text{-}.328)$ & 0.608 $(.599\text{-}.618)$ & 0.913 $(.907\text{-}.919)$\\
        \midrule
        \multirow{7}{*}{\shortstack{5\\($n = 504$)}}
          & CrownGen            & \textbf{30.545} $\mathbf{(30.226\text{-}30.865)}$ & \textbf{50.477} $\mathbf{(49.884\text{-}51.080)}$ & \textbf{0.552} $\mathbf{(.545\text{-}.558)}$ & \textbf{0.847} $\mathbf{(.842\text{-}.852)}$ & \textbf{0.989} $\mathbf{(.988\text{-}.990)}$\\
          & \tiny w/o bound cond. & 41.073 $(40.363\text{-}41.800)$ & 64.051 $(63.021\text{-}65.114)$ & 0.391 $(.383\text{-}.399)$ & 0.686 $(.677\text{-}.695)$ & 0.952 $(.948\text{-}.956)$\\
          & \tiny w/o DITA            & 38.323 $(37.866\text{-}38.796)$ & 76.518 $(75.541\text{-}77.517)$ & 0.420 $(.414\text{-}.426)$ & 0.728 $(.721\text{-}.734)$ & 0.964 $(.961\text{-}.966)$\\
          & \tiny w/o data expansion  & 45.814 $(45.233\text{-}46.405)$ & 82.781 $(81.763\text{-}83.812)$ & 0.332 $(.326\text{-}.338)$ & 0.612 $(.604\text{-}.620)$ & 0.928 $(.924\text{-}.932)$\\
          & PointSea            & 52.088 $(48.869\text{-}55.689)$ & 118.971 $(112.447\text{-}126.135)$ & 0.336 $(.326\text{-}.347)$ & 0.640 $(.626\text{-}.652)$ & 0.928 $(.920\text{-}.936)$\\
          & AdaPoinTr           & 51.450 $(50.018\text{-}53.071)$ & 138.258 $(134.670\text{-}142.006)$ & 0.256 $(.249\text{-}.263)$ & 0.536 $(.526\text{-}.547)$ & 0.892 $(.885\text{-}.898)$\\
          & ProxyFormer         & 48.002 $(47.255\text{-}48.784)$ & 117.438 $(115.676\text{-}119.238)$ & 0.304 $(.298\text{-}.309)$ & 0.598 $(.591\text{-}.606)$ & 0.918 $(.913\text{-}.922)$\\
        \midrule
        \multirow{7}{*}{\shortstack{6\\($n = 496$)}}
          & CrownGen            & \textbf{30.364} $\mathbf{(30.087\text{-}30.650)}$ & \textbf{50.222} $\mathbf{(49.691\text{-}50.771)}$ & \textbf{0.555} $\mathbf{(.550\text{-}.561)}$ & \textbf{0.850} $\mathbf{(.845\text{-}.854)}$ & \textbf{0.989} $\mathbf{(.989\text{-}.990)}$\\
          & \tiny w/o bound cond. & 40.456 $(39.842\text{-}41.102)$ & 63.165 $(62.253\text{-}64.122)$ & 0.398 $(.391\text{-}.405)$ & 0.695 $(.687\text{-}.703)$ & 0.955 $(.951\text{-}.958)$\\
          & \tiny w/o DITA            & 38.322 $(37.912\text{-}38.752)$ & 76.256 $(75.327\text{-}77.212)$ & 0.418 $(.412\text{-}.424)$ & 0.725 $(.719\text{-}.732)$ & 0.965 $(.963\text{-}.967)$\\
          & \tiny w/o data expansion  & 45.130 $(44.597\text{-}45.679)$ & 81.976 $(81.032\text{-}82.956)$ & 0.340 $(.334\text{-}.346)$ & 0.623 $(.615\text{-}.630)$ & 0.931 $(.927\text{-}.935)$\\
          & PointSea            & 45.959 $(43.831\text{-}48.367)$ & 92.886 $(88.795\text{-}97.555)$ & 0.376 $(.368\text{-}.385)$ & 0.673 $(.661\text{-}.684)$ & 0.940 $(.934\text{-}.946)$\\
          & AdaPoinTr           & 70.214 $(60.841\text{-}82.532)$ & 179.146 $(160.437\text{-}203.110)$ & 0.193 $(.186\text{-}.200)$ & 0.451 $(.439\text{-}.463)$ & 0.855 $(.845\text{-}.864)$\\
          & ProxyFormer         & 49.526 $(48.428\text{-}50.699)$ & 121.791 $(118.824\text{-}124.899)$ & 0.285 $(.277\text{-}.293)$ & 0.570 $(.559\text{-}.580)$ & 0.906 $(.899\text{-}.912)$\\
        \bottomrule
 
      \end{tabular}
    \captionsetup{skip=7pt}
      \caption{\textbf{Quantitative evaluation of generated point cloud crowns. (Anteriors)} Mean performance with 95\% CIs (nonparametric bootstrap, $10,000$ resamples) comparing CrownGen with state-of-the-art baselines and ablations, stratified by number of missing anterior (incisors and canines) teeth restored. Metrics marked with $\downarrow$ indicate lower-is-better and vice versa. Chamfer Distance and Earth Mover’s Distance metrics are reported $\times10^3$. Bold denotes the best method.}\label{tab_pointcloud_anterior}
  \end{minipage}
  }
\end{table} 

\begin{table}[ht]
    \centering
    \footnotesize
    \makebox[\textwidth][c]{%
  \begin{minipage}{\textwidth}
    \renewcommand{\arraystretch}{1.2}
      \begin{tabular}{@{}c@{}llll ll @{}}
        \toprule
            \multicolumn{1}{c}{\Centerstack{Number of\\missing teeth\\restored}} 
          &  \multicolumn{1}{c}{} 
          & \multicolumn{1}{c}{\Centerstack{Chamfer\\Distance $L1$ (↓)}} 
          & \multicolumn{1}{c}{\Centerstack{Earth Mover's\\Distance (↓)}} 
          & \multicolumn{1}{c}{$F1$ $@$ 0.3 mm (↑)}
          & \multicolumn{1}{c}{$F1$ $@$ 0.5 mm (↑)}
          & \multicolumn{1}{c}{$F1$ $@$ 1.0 mm (↑)} \\
        \midrule
        \multirow{7}{*}{\shortstack{1\\($n = 1,445$)}}
          & CrownGen            & \textbf{33.921} $\mathbf{(33.586\text{-}34.301)}$ & \textbf{53.174} $\mathbf{(52.665\text{-}53.730)}$ & \textbf{0.474} $\mathbf{(.468\text{-}.479)}$ & \textbf{0.801} $\mathbf{(.796\text{-}.806)}$ & \textbf{0.985} $\mathbf{(.983\text{-}.986)}$\\
          & \tiny w/o bound cond. & 44.457 $(43.818\text{-}45.137)$ & 66.480 $(65.601\text{-}67.414)$ & 0.338 $(.331\text{-}.344)$ & 0.632 $(.624\text{-}.641)$ & 0.940 $(.936\text{-}.944)$\\
          & \tiny w/o DITA            & 40.642 $(40.213\text{-}41.085)$ & 76.214 $(75.332\text{-}77.124)$ & 0.370 $(.364\text{-}.377)$ & 0.691 $(.684\text{-}.698)$ & 0.961 $(.959\text{-}.964)$\\
          & \tiny w/o data expansion  & 47.523 $(46.929\text{-}48.141)$ & 77.695 $(76.825\text{-}78.598)$ & 0.303 $(.297\text{-}.309)$ & 0.588 $(.580\text{-}.597)$ & 0.922 $(.917\text{-}.926)$\\
          & PointSea            & 35.478 $(33.988\text{-}37.950)$ & 55.842 $(53.973\text{-}58.662)$ & 0.471 $(.464\text{-}.479)$ & 0.792 $(.784\text{-}.800)$ & 0.982 $(.978\text{-}.985)$\\
          & AdaPoinTr           & 37.261 $(36.713\text{-}37.861)$ & 59.724 $(58.859\text{-}60.652)$ & 0.435 $(.428\text{-}.443)$ & 0.738 $(.730\text{-}.746)$ & 0.973 $(.970\text{-}.975)$\\
          & ProxyFormer         & 36.495 $(35.899\text{-}37.170)$ & 59.392 $(58.358\text{-}60.533)$ & 0.445 $(.438\text{-}.452)$ & 0.757 $(.750\text{-}.765)$ & 0.974 $(.970\text{-}.977)$\\
        \midrule
        \multirow{7}{*}{\shortstack{2\\($n = 1,618$)}}
          & CrownGen            & \textbf{33.945} $\mathbf{(33.709\text{-}34.193)}$ & \textbf{53.074} $\mathbf{(52.700\text{-}53.459)}$ & \textbf{0.474} $\mathbf{(.470\text{-}.478)}$ & \textbf{0.799} $\mathbf{(.796\text{-}.803)}$ & \textbf{0.985} $\mathbf{(.984\text{-}.986)}$\\
          & \tiny w/o bound cond. & 45.145 $(44.585\text{-}45.766)$ & 67.759 $(66.970\text{-}68.626)$ & 0.335 $(.331\text{-}.340)$ & 0.629 $(.622\text{-}.635)$ & 0.935 $(.932\text{-}.939)$\\
          & \tiny w/o DITA            & 40.520 $(40.217\text{-}40.832)$ & 76.425 $(75.777\text{-}77.089)$ & 0.374 $(.370\text{-}.378)$ & 0.695 $(.690\text{-}.700)$ & 0.961 $(.959\text{-}.962)$\\
          & \tiny w/o data expansion  & 46.919 $(46.507\text{-}47.343)$ & 78.128 $(77.469\text{-}78.812)$ & 0.312 $(.309\text{-}.317)$ & 0.599 $(.593\text{-}.605)$ & 0.924 $(.921\text{-}.927)$\\
          & PointSea            & 51.879 $(48.668\text{-}55.635)$ & 121.209 $(117.115\text{-}126.060)$ & 0.312 $(.306\text{-}.317)$ & 0.619 $(.611\text{-}.627)$ & 0.927 $(.921\text{-}.932)$\\
          & AdaPoinTr           & 47.042 $(45.154\text{-}48.906)$ & 112.841 $(109.411\text{-}116.331)$ & 0.377 $(.370\text{-}.384)$ & 0.671 $(.660\text{-}.682)$ & 0.966 $(.952\text{-}.980)$\\
          & ProxyFormer         & 47.557 $(45.710\text{-}50.320)$ & 91.803 $(88.770\text{-}96.877)$ & 0.325 $(.321\text{-}.329)$ & 0.620 $(.615\text{-}.626)$ & 0.933 $(.930\text{-}.937)$\\
        \midrule
        \multirow{7}{*}{\shortstack{3\\($n = 1,023$)}}
          & CrownGen            & \textbf{33.450} $\mathbf{(33.191\text{-}33.716)}$ & \textbf{52.442} $\mathbf{(52.000\text{-}52.888)}$ & \textbf{0.484} $\mathbf{(.479\text{-}.488)}$ & \textbf{0.807} $\mathbf{(.803\text{-}.811)}$ & \textbf{0.986} $\mathbf{(.985\text{-}.987)}$\\
          & \tiny w/o bound cond. & 44.057 $(43.554\text{-}44.571)$ & 66.481 $(65.736\text{-}67.236)$ & 0.346 $(.341\text{-}.351)$ & 0.642 $(.635\text{-}.648)$ & 0.941 $(.938\text{-}.944)$\\
          & \tiny w/o DITA            & 39.696 $(39.369\text{-}40.018)$ & 75.256 $(74.490\text{-}76.014)$ & 0.387 $(.382\text{-}.391)$ & 0.708 $(.703\text{-}.713)$ & 0.964 $(.962\text{-}.965)$\\
          & \tiny w/o data expansion  & 46.018 $(45.586\text{-}46.444)$ & 77.209 $(76.472\text{-}77.940)$ & 0.323 $(.319\text{-}.327)$ & 0.611 $(.605\text{-}.617)$ & 0.929 $(.926\text{-}.932)$\\
          & PointSea            & 53.291 $(49.264\text{-}58.010)$ & 134.905 $(127.332\text{-}143.939)$ & 0.295 $(.288\text{-}.301)$ & 0.600 $(.590\text{-}.609)$ & 0.922 $(.916\text{-}.928)$\\
          & AdaPoinTr           & 50.099 $(48.688\text{-}52.142)$ & 137.185 $(134.678\text{-}140.301)$ & 0.263 $(.258\text{-}.269)$ & 0.547 $(.539\text{-}.554)$ & 0.906 $(.901\text{-}.910)$\\
          & ProxyFormer         & 52.087 $(51.087\text{-}53.213)$ & 119.931 $(118.044\text{-}121.912)$ & 0.271 $(.266\text{-}.276)$ & 0.542 $(.535\text{-}.549)$ & 0.899 $(.894\text{-}.904)$\\
        \midrule
        \multirow{7}{*}{\shortstack{4\\($n = 995$)}}
          & CrownGen            & \textbf{33.137} $\mathbf{(32.903\text{-}33.380)}$ & \textbf{51.988} $\mathbf{(51.586\text{-}52.396)}$ & \textbf{0.490} $\mathbf{(.486\text{-}.494)}$ & \textbf{0.812} $\mathbf{(.808\text{-}.816)}$ & \textbf{0.986} $\mathbf{(.986\text{-}.987)}$\\
          & \tiny w/o bound cond. & 43.851 $(43.340\text{-}44.396)$ & 66.139 $(65.404\text{-}66.904)$ & 0.352 $(.348\text{-}.357)$ & 0.646 $(.640\text{-}.652)$ & 0.941 $(.938\text{-}.944)$\\
          & \tiny w/o DITA            & 39.435 $(39.133\text{-}39.733)$ & 75.177 $(74.440\text{-}75.905)$ & 0.392 $(.388\text{-}.396)$ & 0.713 $(.708\text{-}.717)$ & 0.964 $(.963\text{-}.966)$\\
          & \tiny w/o data expansion  & 45.488 $(45.098\text{-}45.872)$ & 76.287 $(75.605\text{-}76.964)$ & 0.331 $(.327\text{-}.335)$ & 0.619 $(.614\text{-}.624)$ & 0.932 $(.929\text{-}.934)$\\
          & PointSea            & 48.906 $(46.823\text{-}51.486)$ & 136.551 $(132.507\text{-}141.531)$ & 0.289 $(.283\text{-}.294)$ & 0.599 $(.591\text{-}.608)$ & 0.928 $(.922\text{-}.933)$\\
          & AdaPoinTr           & 48.209 $(45.174\text{-}52.011)$ & 130.666 $(124.671\text{-}138.108)$ & 0.308 $(.302\text{-}.314)$ & 0.605 $(.597\text{-}.612)$ & 0.934 $(.929\text{-}.939)$\\
          & ProxyFormer         & 50.727 $(49.439\text{-}52.598)$ & 129.306 $(126.184\text{-}133.302)$ & 0.263 $(.258\text{-}.268)$ & 0.570 $(.562\text{-}.578)$ & 0.905 $(.900\text{-}.910)$\\
        \midrule
        \multirow{7}{*}{\shortstack{5\\($n = 496$)}}
          & CrownGen            & \textbf{32.737} $\mathbf{(32.405\text{-}33.081)}$ & \textbf{51.444} $\mathbf{(50.875\text{-}52.032)}$ & \textbf{0.500} $\mathbf{(.494\text{-}.505)}$ & \textbf{0.820} $\mathbf{(.815\text{-}.824)}$ & \textbf{0.987} $\mathbf{(.986\text{-}.988)}$\\
          & \tiny w/o bound cond. & 43.039 $(42.313\text{-}43.851)$ & 65.174 $(64.179\text{-}66.208)$ & 0.365 $(.359\text{-}.371)$ & 0.659 $(.652\text{-}.667)$ & 0.944 $(.940\text{-}.948)$\\
          & \tiny w/o DITA            & 38.669 $(38.283\text{-}39.065)$ & 73.759 $(72.769\text{-}74.779)$ & 0.403 $(.397\text{-}.408)$ & 0.725 $(.719\text{-}.731)$ & 0.968 $(.966\text{-}.969)$\\
          & \tiny w/o data expansion  & 44.705 $(44.231\text{-}45.190)$ & 75.590 $(74.681\text{-}76.534)$ & 0.340 $(.335\text{-}.345)$ & 0.630 $(.624\text{-}.637)$ & 0.936 $(.932\text{-}.939)$\\
          & PointSea            & 52.401 $(49.611\text{-}56.043)$ & 147.492 $(142.097\text{-}154.296)$ & 0.264 $(.256\text{-}.272)$ & 0.562 $(.551\text{-}.574)$ & 0.908 $(.900\text{-}.915)$\\
          & AdaPoinTr           & 48.995 $(47.957\text{-}50.091)$ & 155.397 $(152.508\text{-}158.333)$ & 0.245 $(.237\text{-}.253)$ & 0.559 $(.546\text{-}.571)$ & 0.921 $(.914\text{-}.926)$\\
          & ProxyFormer         & 51.662 $(50.899\text{-}52.505)$ & 133.973 $(132.249\text{-}135.785)$ & 0.248 $(.244\text{-}.253)$ & 0.534 $(.527\text{-}.541)$ & 0.903 $(.898\text{-}.907)$\\
        \midrule
        \multirow{7}{*}{\shortstack{6\\($n = 496$)}}
          & CrownGen            & \textbf{32.378} $\mathbf{(32.094\text{-}32.669)}$ & \textbf{50.940} $\mathbf{(50.433\text{-}51.460)}$ & \textbf{0.506} $\mathbf{(.500\text{-}.511)}$ & \textbf{0.824} $\mathbf{(.819\text{-}.828)}$ & \textbf{0.988} $\mathbf{(.987\text{-}.989)}$\\
          & \tiny w/o bound cond. & 42.288 $(41.701\text{-}42.906)$ & 64.221 $(63.370\text{-}65.094)$ & 0.375 $(.369\text{-}.381)$ & 0.669 $(.662\text{-}.676)$ & 0.947 $(.943\text{-}.950)$\\
          & \tiny w/o DITA            & 38.290 $(37.933\text{-}38.649)$ & 73.380 $(72.465\text{-}74.314)$ & 0.410 $(.404\text{-}.415)$ & 0.731 $(.725\text{-}.736)$ & 0.968 $(.967\text{-}.970)$\\
          & \tiny w/o data expansion  & 44.020 $(43.587\text{-}44.460)$ & 74.813 $(73.967\text{-}75.690)$ & 0.349 $(.344\text{-}.354)$ & 0.641 $(.635\text{-}.647)$ & 0.938 $(.935\text{-}.941)$\\
          & PointSea            & 45.874 $(44.073\text{-}48.395)$ & 117.662 $(114.473\text{-}122.220)$ & 0.332 $(.324\text{-}.340)$ & 0.645 $(.634\text{-}.656)$ & 0.936 $(.930\text{-}.942)$\\
          & AdaPoinTr           & 51.607 $(48.673\text{-}55.897)$ & 150.636 $(144.542\text{-}159.103)$ & 0.243 $(.237\text{-}.250)$ & 0.553 $(.543\text{-}.563)$ & 0.908 $(.902\text{-}.913)$\\
          & ProxyFormer         & 52.001 $(50.924\text{-}53.312)$ & 140.488 $(137.850\text{-}143.352)$ & 0.237 $(.231\text{-}.243)$ & 0.535 $(.526\text{-}.544)$ & 0.898 $(.891\text{-}.904)$\\
        \bottomrule
 
      \end{tabular}
    \captionsetup{skip=7pt}
      \caption{\textbf{Quantitative evaluation of generated point cloud crowns. (Premolars)} Mean performance with 95\% CIs (nonparametric bootstrap, $10,000$ resamples) comparing CrownGen with state-of-the-art baselines and ablations, stratified by number of missing premolar teeth restored. Metrics marked with $\downarrow$ indicate lower-is-better and vice versa. Chamfer Distance and Earth Mover’s Distance metrics are reported $\times10^3$. Bold denotes the best method.}\label{tab_pointcloud_premolar}
  \end{minipage}
  }
\end{table} 

\begin{table}[ht]
    \centering
    \footnotesize
    \makebox[\textwidth][c]{%
  \begin{minipage}{\textwidth}
  \renewcommand{\arraystretch}{1.2}
      \begin{tabular}{@{}c@{}llll ll @{}}
        \toprule
            \multicolumn{1}{c}{\Centerstack{Number of\\missing teeth\\restored}} 
          &  \multicolumn{1}{c}{} 
          & \multicolumn{1}{c}{\Centerstack{Chamfer\\Distance $L1$ (↓)}} 
          & \multicolumn{1}{c}{\Centerstack{Earth Mover's\\Distance (↓)}} 
          & \multicolumn{1}{c}{$F1$ $@$ 0.3 mm (↑)}
          & \multicolumn{1}{c}{$F1$ $@$ 0.5 mm (↑)}
          & \multicolumn{1}{c}{$F1$ $@$ 1.0 mm (↑)} \\
        \midrule
        \multirow{7}{*}{\shortstack{1\\($n = 1,422$)}}
          & CrownGen            & \textbf{40.981} $\mathbf{(40.655\text{-}41.316)}$ & \textbf{63.369} $\mathbf{(62.827\text{-}63.921)}$ & \textbf{0.352} $\mathbf{(.348\text{-}.356)}$ & \textbf{0.694} $\mathbf{(.689\text{-}.699)}$ & \textbf{0.963} $\mathbf{(.961\text{-}.965)}$\\
          & \tiny w/o bound cond. & 58.691 $(57.725\text{-}59.700)$ & 87.718 $(86.359\text{-}89.129)$ & 0.221 $(.216\text{-}.226)$ & 0.485 $(.476\text{-}.494)$ & 0.844 $(.836\text{-}.852)$\\
          & \tiny w/o DITA            & 48.750 $(48.229\text{-}49.290)$ & 86.872 $(85.890\text{-}87.894)$ & 0.268 $(.263\text{-}.272)$ & 0.580 $(.573\text{-}.587)$ & 0.923 $(.919\text{-}.927)$\\
          & \tiny w/o data expansion  & 55.920 $(55.188\text{-}56.675)$ & 88.750 $(87.757\text{-}89.768)$ & 0.231 $(.227\text{-}.235)$ & 0.502 $(.495\text{-}.510)$ & 0.865 $(.859\text{-}.871)$\\
          & PointSea            & 51.609 $(45.359\text{-}59.661)$ & 79.290 $(70.122\text{-}91.313)$ & 0.346 $(.340\text{-}.352)$ & 0.673 $(.664\text{-}.682)$ & 0.943 $(.936\text{-}.949)$\\
          & AdaPoinTr           & 46.815 $(44.796\text{-}49.633)$ & 77.626 $(75.404\text{-}80.588)$ & 0.323 $(.318\text{-}.329)$ & 0.640 $(.631\text{-}.648)$ & 0.933 $(.928\text{-}.937)$\\
          & ProxyFormer         & 49.889 $(46.349\text{-}55.083)$ & 77.394 $(73.202\text{-}83.389)$ & 0.317 $(.311\text{-}.322)$ & 0.629 $(.621\text{-}.637)$ & 0.923 $(.917\text{-}.929)$\\
        \midrule
        \multirow{7}{*}{\shortstack{2\\($n = 1,660$)}}
          & CrownGen            & \textbf{41.312} $\mathbf{(41.057\text{-}41.576)}$ & \textbf{63.861} $\mathbf{(63.450\text{-}64.285)}$ & \textbf{0.351} $\mathbf{(.347\text{-}.354)}$ & \textbf{0.690} $\mathbf{(.686\text{-}.694)}$ & \textbf{0.961} $\mathbf{(.959\text{-}.962)}$\\
          & \tiny w/o bound cond. & 58.147 $(57.490\text{-}58.834)$ & 87.433 $(86.502\text{-}88.407)$ & 0.230 $(.227\text{-}.234)$ & 0.496 $(.490\text{-}.502)$ & 0.846 $(.841\text{-}.851)$\\
          & \tiny w/o DITA            & 48.177 $(47.823\text{-}48.546)$ & 85.839 $(85.159\text{-}86.548)$ & 0.277 $(.274\text{-}.280)$ & 0.591 $(.586\text{-}.596)$ & 0.925 $(.922\text{-}.927)$\\
          & \tiny w/o data expansion  & 55.952 $(55.448\text{-}56.474)$ & 89.225 $(88.522\text{-}89.955)$ & 0.233 $(.230\text{-}.236)$ & 0.503 $(.498\text{-}.508)$ & 0.864 $(.860\text{-}.868)$\\
          & PointSea            & 73.502 $(65.581\text{-}82.818)$ & 162.564 $(152.032\text{-}174.682)$ & 0.216 $(.211\text{-}.220)$ & 0.485 $(.478\text{-}.493)$ & 0.851 $(.844\text{-}.857)$\\
          & AdaPoinTr           & 58.101 $(55.696\text{-}60.536)$ & 153.497 $(148.657\text{-}158.428)$ & 0.212 $(.208\text{-}.216)$ & 0.513 $(.505\text{-}.521)$ & 0.837 $(.826\text{-}.847)$\\
          & ProxyFormer         & 69.722 $(67.612\text{-}72.177)$ & 132.323 $(128.603\text{-}136.555)$ & 0.205 $(.201\text{-}.209)$ & 0.432 $(.426\text{-}.439)$ & 0.797 $(.791\text{-}.803)$\\
        \midrule
        \multirow{7}{*}{\shortstack{3\\($n = 1,025$)}}
          & CrownGen            & \textbf{40.586} $\mathbf{(40.303\text{-}40.865)}$ & \textbf{62.844} $\mathbf{(62.386\text{-}63.296)}$ & \textbf{0.362} $\mathbf{(.359\text{-}.366)}$ & \textbf{0.701} $\mathbf{(.696\text{-}.705)}$ & \textbf{0.964} $\mathbf{(.962\text{-}.965)}$\\
          & \tiny w/o bound cond. & 56.411 $(55.719\text{-}57.107)$ & 85.828 $(84.786\text{-}86.862)$ & 0.244 $(.240\text{-}.248)$ & 0.515 $(.509\text{-}.521)$ & 0.859 $(.854\text{-}.864)$\\
          & \tiny w/o DITA            & 47.077 $(46.685\text{-}47.462)$ & 84.551 $(83.764\text{-}85.326)$ & 0.290 $(.287\text{-}.294)$ & 0.607 $(.602\text{-}.612)$ & 0.930 $(.927\text{-}.933)$\\
          & \tiny w/o data expansion  & 54.641 $(54.110\text{-}55.161)$ & 88.012 $(87.233\text{-}88.767)$ & 0.245 $(.241\text{-}.248)$ & 0.519 $(.513\text{-}.524)$ & 0.873 $(.869\text{-}.877)$\\
          & PointSea            & 64.832 $(59.956\text{-}70.901)$ & 162.158 $(152.803\text{-}173.892)$ & 0.208 $(.203\text{-}.213)$ & 0.484 $(.474\text{-}.492)$ & 0.857 $(.848\text{-}.864)$\\
          & AdaPoinTr           & 63.888 $(61.245\text{-}67.245)$ & 167.936 $(163.680\text{-}173.146)$ & 0.191 $(.187\text{-}.195)$ & 0.447 $(.439\text{-}.454)$ & 0.817 $(.811\text{-}.824)$\\
          & ProxyFormer         & 74.305 $(69.909\text{-}80.452)$ & 166.772 $(158.374\text{-}178.738)$ & 0.186 $(.182\text{-}.191)$ & 0.397 $(.390\text{-}.404)$ & 0.777 $(.768\text{-}.785)$\\
        \midrule
        \multirow{7}{*}{\shortstack{4\\($n = 997$)}}
          & CrownGen            & \textbf{40.186} $\mathbf{(39.929\text{-}40.443)}$ & \textbf{62.394} $\mathbf{(61.974\text{-}62.820)}$ & \textbf{0.370} $\mathbf{(.366\text{-}.373)}$ & \textbf{0.708} $\mathbf{(.704\text{-}.712)}$ & \textbf{0.964} $\mathbf{(.963\text{-}.966)}$\\
          & \tiny w/o bound cond. & 55.004 $(54.415\text{-}55.596)$ & 83.949 $(83.052\text{-}84.851)$ & 0.255 $(.252\text{-}.259)$ & 0.531 $(.525\text{-}.536)$ & 0.868 $(.863\text{-}.872)$\\
          & \tiny w/o DITA            & 46.503 $(46.153\text{-}46.853)$ & 83.896 $(83.122\text{-}84.666)$ & 0.299 $(.295\text{-}.302)$ & 0.616 $(.612\text{-}.621)$ & 0.933 $(.930\text{-}.935)$\\
          & \tiny w/o data expansion  & 53.996 $(53.524\text{-}54.465)$ & 87.306 $(86.583\text{-}88.034)$ & 0.253 $(.250\text{-}.257)$ & 0.529 $(.524\text{-}.534)$ & 0.876 $(.872\text{-}.880)$\\
          & PointSea            & 66.256 $(62.029\text{-}71.495)$ & 173.557 $(165.633\text{-}183.542)$ & 0.192 $(.187\text{-}.197)$ & 0.444 $(.435\text{-}.452)$ & 0.835 $(.828\text{-}.843)$\\
          & AdaPoinTr           & 58.156 $(56.723\text{-}59.968)$ & 155.547 $(152.425\text{-}159.277)$ & 0.216 $(.211\text{-}.220)$ & 0.487 $(.479\text{-}.495)$ & 0.845 $(.839\text{-}.851)$\\
          & ProxyFormer         & 69.250 $(65.799\text{-}73.939)$ & 177.049 $(170.796\text{-}185.773)$ & 0.173 $(.168\text{-}.179)$ & 0.423 $(.413\text{-}.433)$ & 0.814 $(.803\text{-}.824)$\\
        \midrule
        \multirow{7}{*}{\shortstack{5\\($n = 497$)}}
          & CrownGen            & \textbf{39.749} $\mathbf{(39.393\text{-}40.120)}$ & \textbf{61.775} $\mathbf{(61.214\text{-}62.364)}$ & \textbf{0.378} $\mathbf{(.373\text{-}.383)}$ & \textbf{0.715} $\mathbf{(.710\text{-}.721)}$ & \textbf{0.965} $\mathbf{(.964\text{-}.967)}$\\
          & \tiny w/o bound cond. & 53.946 $(53.204\text{-}54.722)$ & 82.502 $(81.391\text{-}83.650)$ & 0.266 $(.261\text{-}.271)$ & 0.544 $(.537\text{-}.551)$ & 0.874 $(.869\text{-}.880)$\\
           & \tiny w/o DITA            & 45.786 $(45.312\text{-}46.275)$ & 82.631 $(81.583\text{-}83.711)$ & 0.308 $(.303\text{-}.313)$ & 0.627 $(.621\text{-}.633)$ & 0.936 $(.933\text{-}.939)$\\
           & \tiny w/o data expansion  & 52.524 $(51.896\text{-}53.170)$ & 85.808 $(84.833\text{-}86.811)$ & 0.266 $(.261\text{-}.271)$ & 0.547 $(.541\text{-}.554)$ & 0.886 $(.881\text{-}.890)$\\
          & PointSea            & 72.755 $(67.188\text{-}80.364)$ & 200.138 $(189.642\text{-}214.753)$ & 0.141 $(.135\text{-}.147)$ & 0.359 $(.347\text{-}.370)$ & 0.769 $(.755\text{-}.782)$\\
          & AdaPoinTr           & 61.980 $(60.722\text{-}63.327)$ & 171.821 $(168.840\text{-}174.907)$ & 0.159 $(.154\text{-}.165)$ & 0.411 $(.401\text{-}.421)$ & 0.821 $(.813\text{-}.829)$\\
          & ProxyFormer         & 98.573 $(87.254\text{-}111.077)$ & 237.293 $(215.636\text{-}261.529)$ & 0.157 $(.153\text{-}.161)$ & 0.363 $(.355\text{-}.371)$ & 0.758 $(.747\text{-}.769)$\\
        \midrule
        \multirow{7}{*}{\shortstack{6\\($n = 496$)}}
          & CrownGen            & \textbf{39.084} $\mathbf{(38.758\text{-}39.418)}$ & \textbf{60.878} $\mathbf{(60.336\text{-}61.433)}$ & \textbf{0.387} $\mathbf{(.383\text{-}.392)}$ & \textbf{0.725} $\mathbf{(.720\text{-}.730)}$ & \textbf{0.968} $\mathbf{(.967\text{-}.970)}$\\
          & \tiny w/o bound cond. & 52.864 $(52.235\text{-}53.520)$ & 81.722 $(80.695\text{-}82.793)$ & 0.277 $(.272\text{-}.281)$ & 0.558 $(.552\text{-}.564)$ & 0.880 $(.876\text{-}.88\tiny 5)$\\
          & \tiny w/o DITA            & 45.043 $(44.633\text{-}45.464)$ & 81.988 $(81.026\text{-}82.974)$ & 0.317 $(.313\text{-}.321)$ & 0.638 $(.633\text{-}.643)$ & 0.940 $(.937\text{-}.9\tiny 42)$\\
          & \tiny w/o data expansion  & 51.913 $(51.364\text{-}52.483)$ & 84.953 $(84.084\text{-}85.848)$ & 0.273 $(.269\text{-}.277)$ & 0.556 $(.550\text{-}.562)$ & 0.889 $(.885\text{-}.893)$\\
          & PointSea            & 82.317 $(73.255\text{-}93.130)$ & 196.359 $(179.529\text{-}216.543)$ & 0.182 $(.176\text{-}.189)$ & 0.429 $(.416\text{-}.441)$ & 0.788 $(.774\text{-}.802)$\\
          & AdaPoinTr           & 67.284 $(65.295\text{-}69.505)$ & 171.985 $(168.354\text{-}175.779)$ & 0.163 $(.157\text{-}.168)$ & 0.405 $(.395\text{-}.415)$ & 0.796 $(.788\text{-}.804)$\\
          & ProxyFormer         & 135.775 $(120.517\text{-}151.695)$ & 320.356 $(291.166\text{-}350.989)$ & 0.122 $(.116\text{-}.127)$ & 0.309 $(.296\text{-}.322)$ & 0.680 $(.657\text{-}.703)$\\
        \bottomrule
 
      \end{tabular}
    \captionsetup{skip=7pt}
      \caption{\textbf{Quantitative evaluation of generated point cloud crowns. (Molars)} Mean performance with 95\% CIs (nonparametric bootstrap, $10,000$ resamples) comparing CrownGen with state-of-the-art baselines and ablations, stratified by number of missing molar teeth restored. Metrics marked with $\downarrow$ indicate lower-is-better and vice versa. Chamfer Distance and Earth Mover’s Distance metrics are reported $\times10^3$. Bold denotes the best method.}\label{tab_pointcloud_molar}
  \end{minipage}
  }
\end{table} 

\begin{sidewaystable}
\setlength{\tabcolsep}{4pt}
  \renewcommand{\arraystretch}{2.0}
\begin{tabular*}{\textwidth}{@{\extracolsep\fill}lcccccccc}
\toprule%
& \multicolumn{4}{@{}c}{Normal Consistency ($\uparrow$)}
& \multicolumn{4}{c@{}}{Average Surface Distance ($\downarrow$)} \\
\cmidrule(l){2-5} \cmidrule(r){6-9}
Tooth type&CrownGen&\Centerstack{w/o bound\\condition}&\Centerstack{w/o\\DITA} & \Centerstack{w/o training\\data expansion}&CrownGen&\Centerstack{w/o bound\\condition}&\Centerstack{w/o\\DITA} & \Centerstack{w/o training\\data expansion}\\
\midrule
\Centerstack[l]{All types\\$(n=16,368)$}          & \Centerstack{\textbf{0.925}\\$(\mathbf{0.924\text{-}0.925)}$} & \Centerstack{0.887\\$(0.886\text{-}0.888)$} & \Centerstack{0.915\\$(0.914\text{-}0.916)$} & \Centerstack{0.884\\$(0.883\text{-}0.885)$} & \Centerstack{\textbf{0.267}\\$\mathbf{(0.266\text{-}0.269)}$} & \Centerstack{0.411\\$(0.408\text{-}0.414)$} & \Centerstack{0.333\\$(0.332\text{-}0.335)$} & \Centerstack{0.438\\$(0.435\text{-}0.440)$}\\
\Centerstack[l]{Central incisor\\$(n=2,347)$}     & \Centerstack{\textbf{0.933}\\$(\mathbf{0.932\text{-}0.934)}$} & \Centerstack{0.900\\$(0.897\text{-}0.902)$} & \Centerstack{0.928\\$(0.927\text{-}0.930)$} & \Centerstack{0.893\\$(0.890\text{-}0.896)$} & \Centerstack{\textbf{0.213}\\$\mathbf{(0.210\text{-}0.215)}$} & \Centerstack{0.327\\$(0.321\text{-}0.333)$} & \Centerstack{0.296\\$(0.291\text{-}0.301)$} & \Centerstack{0.389\\$(0.382\text{-}0.396)$}\\
\Centerstack[l]{Lateral incisor\\$(n=2,277)$}     & \Centerstack{\textbf{0.929}\\$(\mathbf{0.928\text{-}0.930)}$} & \Centerstack{0.891\\$(0.889\text{-}0.894)$} & \Centerstack{0.921\\$(0.919\text{-}0.923)$} & \Centerstack{0.885\\$(0.882\text{-}0.888)$} & \Centerstack{\textbf{0.226}\\$\mathbf{(0.223\text{-}0.228)}$} & \Centerstack{0.348\\$(0.342\text{-}0.354)$} & \Centerstack{0.303\\$(0.299\text{-}0.308)$} & \Centerstack{0.395\\$(0.388\text{-}0.402)$}\\
\Centerstack[l]{Canine\\$(n=2,392)$}              & \Centerstack{\textbf{0.928}\\$(\mathbf{0.927\text{-}0.930)}$} & \Centerstack{0.889\\$(0.887\text{-}0.892)$} & \Centerstack{0.919\\$(0.917\text{-}0.921)$} & \Centerstack{0.899\\$(0.897\text{-}0.901)$} & \Centerstack{\textbf{0.267}\\$\mathbf{(0.264\text{-}0.270)}$} & \Centerstack{0.405\\$(0.398\text{-}0.413)$} & \Centerstack{0.343\\$(0.338\text{-}0.348)$} & \Centerstack{0.446\\$(0.439\text{-}0.453)$}\\
\Centerstack[l]{First premolar\\$(n=2,335)$}      & \Centerstack{\textbf{0.928}\\$(\mathbf{0.927\text{-}0.929)}$} & \Centerstack{0.898\\$(0.896\text{-}0.900)$} & \Centerstack{0.917\\$(0.915\text{-}0.918)$} & \Centerstack{0.886\\$(0.884\text{-}0.888)$} & \Centerstack{\textbf{0.265}\\$\mathbf{(0.261\text{-}0.268)}$} & \Centerstack{0.397\\$(0.386\text{-}0.409)$} & \Centerstack{0.319\\$(0.315\text{-}0.322)$} & \Centerstack{0.421\\$(0.415\text{-}0.427)$}\\
\Centerstack[l]{Second premolar\\$(n=2,333)$}     & \Centerstack{\textbf{0.928}\\$(\mathbf{0.926\text{-}0.929)}$} & \Centerstack{0.890\\$(0.888\text{-}0.893)$} & \Centerstack{0.916\\$(0.914\text{-}0.917)$} & \Centerstack{0.882\\$(0.880\text{-}0.884)$} & \Centerstack{\textbf{0.258}\\$\mathbf{(0.255\text{-}0.261)}$} & \Centerstack{0.396\\$(0.390\text{-}0.402)$} & \Centerstack{0.313\\$(0.309\text{-}0.316)$} & \Centerstack{0.413\\$(0.407\text{-}0.418)$}\\
\Centerstack[l]{First molar\\$(n=2,355)$}         & \Centerstack{\textbf{0.923}\\$(\mathbf{0.922\text{-}0.923)}$} & \Centerstack{0.893\\$(0.891\text{-}0.894)$} & \Centerstack{0.912\\$(0.911\text{-}0.913)$} & \Centerstack{0.884\\$(0.883\text{-}0.886)$} & \Centerstack{\textbf{0.296}\\$\mathbf{(0.293\text{-}0.298)}$} & \Centerstack{0.434\\$(0.428\text{-}0.440)$} & \Centerstack{0.354\\$(0.350\text{-}0.358)$} & \Centerstack{0.460\\$(0.454\text{-}0.467)$}\\
\Centerstack[l]{Second molar\\$(n=2,329)$}         & \Centerstack{\textbf{0.905}\\$(\mathbf{0.903\text{-}0.906)}$} & \Centerstack{0.845\\$(0.841\text{-}0.848)$} & \Centerstack{0.892\\$(0.890\text{-}0.894)$} & \Centerstack{0.856\\$(0.854\text{-}0.859)$} & \Centerstack{\textbf{0.347}\\$\mathbf{(0.344\text{-}0.351)}$} & \Centerstack{0.572\\$(0.563\text{-}0.581)$} & \Centerstack{0.405\\$(0.400\text{-}0.409)$} & \Centerstack{0.538\\$(0.530\text{-}0.546)$}\\
\botrule
\end{tabular*}
\captionsetup{skip=7pt}
\caption{\textbf{Ablation study on final reconstructed mesh quality.} Mean performance with 95\% CIs (nonparametric bootstrap, $10,000$ resamples) of CrownGen and its three ablated variants on the final reconstructed mesh surfaces across scenarios derived from 496 test dentitions. Metrics marked with $\downarrow$ indicate lower-is-better and vice versa. Bold denotes the best method.}\label{tab_ablation_metric}
\end{sidewaystable} 

\vspace*{\fill}
\begin{table}[hp]
    \centering
      \renewcommand{\arraystretch}{1.5}
    \makebox[0.95\textwidth][c]{%
    \begin{minipage}{0.95\textwidth}

        \begin{tabular}{@{}p{0.15\textwidth} c @{} i{\textwidth}@{}}
               
            \toprule
            \multicolumn{1}{@{}l}{Criteria} &
            \multicolumn{1}{c@{}}{Score} &
            \multicolumn{1}{c@{}}{Description} \\
            \midrule
            \multirow{5}{=}{\raggedright\textbf{Occlusion} \par (Molars, Premolars, Canines)} 
            & 1 & 
              \item Centric occlusion is characterized by cusp-to-cusp contact or a complete lack of occlusal contact.
              \item The design exhibits a collapsed marginal ridge, representing a major morphological failure.
             \\
            \cmidrule(lr){2-3}
            & 2 &
              \item Centric occlusion is established in a non-anatomic cusp-to-ridge relationship.
              \item The marginal ridge displays a visible discontinuity or mismatch with the adjacent tooth.
             \\
            \cmidrule(lr){2-3}
            & 3 & 
            \item Functional cusp tips are correctly centered within the fossae, with any minor deviation limited to the fossa slopes.
              \item The marginal ridge is continuous with adjacent teeth, presenting at most a slight step.
             \\
            \midrule
            \multirow{5.5}{=}{\raggedright\textbf{Occlusion} \par (Incisors)}
            & 1 &           
              \item The lingual anatomy is incorrect and completely blocks anterior guidance.
              \item The incisal edge displays a major deviation from the ideal position.
              \item The design results in an open bite with no functional contact.
             \\
            \cmidrule(lr){2-3}
            & 2 &           
              \item The lingual anatomy is present but impedes proper anterior guidance.
              \item The incisal edge displays a minor misalignment.
             \\
            \cmidrule(lr){2-3}
            & 3 &           
              \item Centric contact is stable and light, or there is no contact, with only minor deviations in the lingual contour.
              \item Overjet and overbite are within normal anatomical limits.
              \item The incisal edge is properly aligned both vertically and horizontally.
             \\
            \midrule
        
            \multirow{5.5}{=}{\raggedright\textbf{Proximal contact}}
            & 1 &           
              \item The proximal contact is absent (open contact) or grossly overlapping.
              \item The contact point is misplaced relative to its correct anatomical position.
             \\
            \cmidrule(lr){2-3}
            & 2 &           
              \item The proximal contact is correctly located but exhibits an inadequate size or shape.
              \item The surrounding embrasure form is incorrect or non-anatomical.
             \\
            \cmidrule(lr){2-3}
            & 3 &           
              \item The proximal contact is located in the correct anatomical position.
              \item The contact's size and shape are appropriate, ranging from a point to a fully anatomical surface.
              \item The embrasure form is properly defined and anatomical.
             \\
            \midrule
        
            \multirow{5}{=}{\raggedright\textbf{Alignment with arch form}}
            & 1 &           
              \item The crown is grossly malaligned, appearing significantly buccal, lingual, or recessed.
              \item The design exhibits a severe rotation that disrupts the natural arch curvature.
             \\
            \cmidrule(lr){2-3}
            & 2 &           
              \item The crown is in a noticeable malposition relative to the arch form.
              \item Correcting the malposition would require significant morphological alteration to the crown.
             \\
            \cmidrule(lr){2-3}
            & 3 &           
              \item The crown integrates into a smooth, continuous arch curve with at most a slight deviation.
              \item No protrusion or recession relative to adjacent teeth; any discrepancy is correctable with minimal recontouring.
             \\
            \midrule
        
            \multirow{5}{=}{\raggedright\textbf{Crown form \& contour}}
            & 1 &           
              \item The crown exhibits clear and significant over- or under-contouring.
              \item The embrasure spaces are poorly defined.
              \item The overall shape is fundamentally non-anatomical.
             \\
            \cmidrule(lr){2-3}
            & 2 &           
              \item The crown silhouette is recognizable but displays moderate contour deviations.
              \item The embrasures are partially collapsed or excessively open.
             \\
            \cmidrule(lr){2-3}
            & 3 &           
              \item The crown silhouette is anatomically correct.
              \item The surface convexities and concavities are proper, with only subtle deviations.
              \item The embrasures are well-formed, displaying only minimal irregularities.
             \\
            \bottomrule
        
          \end{tabular}
          \captionsetup{skip=7pt}
          \caption{\textbf{Digital crown quality rubric.} The digital crown designs were evaluated by two trained dentists using the rubric detailed above. Each of the four primary evaluation criteria is presented on a 3-point scale calibrated to the anticipated downstream manual workload required for clinical viability. A score of 3 represents the highest quality (Clinically acceptable) and 1 represents the lowest quality (Clinically unacceptable).}\label{tab_rubric}

          \end{minipage}
  }
\end{table} 
\vspace*{\fill}

\vspace*{\fill}
\begin{table}[h]
  \centering
  \renewcommand{\arraystretch}{1.2}
  \begin{tabular}{@{} ll ll @{}}
    \toprule
    \multicolumn{2}{c}{} 
      & \multicolumn{1}{c}{Implant-borne} 
      & \multicolumn{1}{c}{Traditional} \\
    \midrule
    \multirow{3}{*}{Maxillary}
      & Anterior & 21            & 11               \\
      & Premolar & 24, 23            & 15, 14           \\
      & Molar    & 26, 16, 17, 27    & 16, 16, 27       \\
    \midrule
    \multirow{3}{*}{Mandibular}
      & Anterior & 41, 31                & 42               \\
      & Premolar & 34                & 45, 45           \\
      & Molar    & 47, 37, 37        & 36, 37, 46, 46   \\
    \bottomrule
  \end{tabular}
    \captionsetup{skip=7pt}
  \caption{\textbf{Reader-study cohort and restoration summary.} Outline of the 26 restoration cases from 23 patients for the reader study. The numbers indicate the tooth type in FDI notation.}
  \label{tab_reader_demographic}
\end{table} 

\begin{table}[h]
  \renewcommand{\arraystretch}{1.5}
\begin{tabular}{lcccc}
\toprule
& \multicolumn{2}{@{}c}{CrownGen-assisted}
& \multicolumn{2}{c@{}}{Fully manual} \\
\cmidrule(l){2-3}\cmidrule(r){4-5}
Endpoint&$\text{Mean}\pm\text{SD}$&95\% CI&$\text{Mean}\pm\text{SD}$ & 95\% CI\\
\midrule
Composite                & $2.938\pm0.194$ & $(2.851-2.995)$ & $2.928\pm0.197$ & $(2.841-2.990)$ \\
Occlusion                & $2.942\pm0.216$ & $(2.846-3.000)$ & $2.904\pm0.246$ & $(2.808-2.981)$ \\
Proximal contact         & $2.942\pm0.163$ & $(2.865-3.000)$ & $2.923\pm0.184$ & $(2.846-2.981)$ \\
Alignment with arch form & $2.942\pm0.216$ & $(2.846-3.000)$ & $2.942\pm0.216$ & $(2.846-3.000)$ \\
Crown form \& contour    & $2.923\pm0.232$ & $(2.827-3.000)$ & $2.942\pm0.216$ & $(2.846-3.000)$ \\

\botrule
\end{tabular}
\captionsetup{skip=7pt}
\caption{\textbf{Case-level average scores by workflow and criteria $\mathbf{(n=26)}$.} For each case, the two readers’ scores were averaged per criterion; the composite is the per-case mean across the four criteria. 95\% CIs are estimated via case-level nonparametric bootstrap with $10,000$ resamples.}\label{tab_reader_summary}
\end{table} 
\vspace*{\fill}

\end{appendices}
\end{document}